\pdfoutput=1

\documentclass{article}

\usepackage[main, final]{neurips_2026}

\usepackage[utf8]{inputenc}
\usepackage[T1]{fontenc}
\usepackage{microtype}
\usepackage{graphicx}
\usepackage[font=small]{caption}
\usepackage{subcaption}
\usepackage{booktabs}
\usepackage{wrapfig}
\usepackage{hyperref}
\hypersetup{
  pdftitle={Evolutionary System Prompt Learning for Reinforcement Learning in LLMs},
  pdfauthor={Lunjun Zhang, Ryan Chen, Bradly C. Stadie}
}

\usepackage{algorithm}
\usepackage{algorithmic}
\usepackage{float}
\usepackage{placeins}

\usepackage{amsmath}
\usepackage{amssymb}
\usepackage{mathtools}
\usepackage{amsthm}

\usepackage{array}
\usepackage[dvipsnames]{xcolor}

\usepackage[capitalize,noabbrev]{cleveref}

\usepackage[most]{tcolorbox}
\usepackage{soul}
\sethlcolor{yellow}
\newtcolorbox{runbox}[1][]{
  enhanced,
  colback=white, colframe=black!20,
  boxrule=0.5pt, arc=1.5mm,
  colbacktitle=black!7, coltitle=black!85, fonttitle=\bfseries,
  left=2mm,right=2mm,top=1mm,bottom=1mm,
  #1
}

\newtcolorbox{runcard}[1][]{
  enhanced,
  colback=white,
  colframe=black!25,
  boxrule=0.5pt,
  arc=1.5mm,
  colbacktitle=black!12, coltitle=black!85,
  left=2mm,right=2mm,top=1mm,bottom=1mm,
  width=\textwidth,
  #1
}

\theoremstyle{plain}

\theoremstyle{definition}

\theoremstyle{remark}

\usepackage[disable,textsize=tiny]{todonotes}

\usepackage{fvextra} %

\DefineVerbatimEnvironment{WrapVerbatim}{Verbatim}{
  breaklines=true,
  breakanywhere=true,
  breaksymbolleft=,
  fontsize=\small
}

\DefineVerbatimEnvironment{WrapVerbatimInner}{Verbatim}{
  breaklines=true,
  breakanywhere=true,
  breaksymbolleft=,
  fontsize=\small
}

\newenvironment{principlelist}{%
  \begin{list}{}{%
    \setlength{\labelwidth}{2.7em}%
    \setlength{\labelsep}{0.5em}%
    \setlength{\leftmargin}{3.2em}%
    \setlength{\itemsep}{3pt}%
    \setlength{\parsep}{0pt}%
    \setlength{\topsep}{2pt}%
    \renewcommand{\makelabel}[1]{##1\hfil}}}%
  {\end{list}}

\DefineVerbatimEnvironment{PromptVerbatim}{Verbatim}{
  breaklines=true,
  breaksymbolleft=,
  fontfamily=\rmdefault,
  fontsize=\normalsize
}

\title{Evolutionary System Prompt Learning\\ for Reinforcement Learning in LLMs}

\author{
  Lunjun Zhang$^{1}$\thanks{Corresponding author: \texttt{lunjun@cs.toronto.edu}} \qquad Ryan Chen$^{2,3}$ \qquad Bradly C. Stadie$^{2,3}$ \\
  $^{1}$University of Toronto \qquad $^{2}$Northwestern University \\
  $^{3}$Bridgewater AIA Labs \\
}

\begin{document}

\setcounter{footnote}{1}
\maketitle
\setcounter{footnote}{0}

\begin{abstract}
Building agentic systems that can autonomously self-improve from experience is a longstanding goal of AI. Large language models (LLMs) today primarily self-improve via two mechanisms: self-reflection for context updates, and reinforcement learning (RL) for weight updates. In this work, we propose \textbf{Evolutionary System Prompt Learning (E-SPL)}, a method for jointly improving a model's experiential context and its weights. In each RL iteration, E-SPL samples trajectories under multiple system prompts in parallel, then uses them to jointly update the weights with RL and evolve the system prompts with self-reflection. E-SPL encourages a natural division between procedural knowledge in weights and interpretable declarative knowledge in system prompts. Most notably, in an easy-to-hard generalization setting (AIME $\rightarrow$ BeyondAIME), E-SPL improves RL success rate from $38.8\% \rightarrow 45.1\%$. E-SPL also improves RL on AIME 2025 ($56.3\% \rightarrow 60.6\%$), HMMT 2025 ($50.0\% \rightarrow 52.7\%$), and agentic search ($44.2\% \rightarrow 48.6\%$ on gpt-oss-120b). Across all settings, E-SPL outperforms both RL-only and evolution-only baselines, demonstrating that weight updates and context updates are deeply synergistic and can together yield gains in generalization that neither achieves alone. Code is available at \url{https://github.com/LunjunZhang/E-SPL}.
\end{abstract}

\section{Introduction}

\begin{wrapfigure}{r}{0.5\textwidth}
  \vspace{-12pt}
  \centering
  \includegraphics[width=0.49\textwidth]{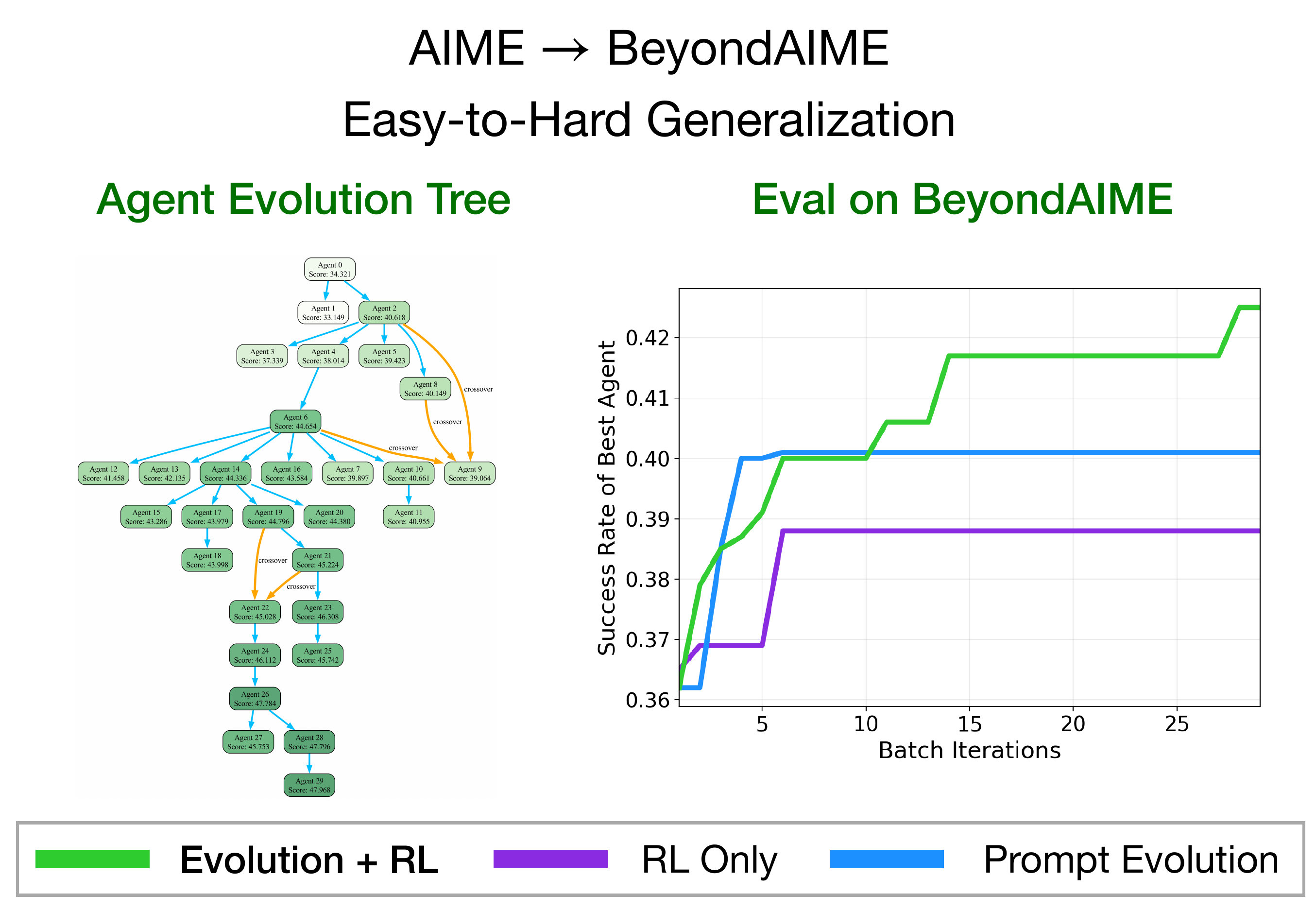}
  \caption{\textbf{Evolving System Prompts helps RL generalize better}.}
  \label{fig:hook}

  \vspace{-12pt}
\end{wrapfigure}

Evolutionary algorithms (EA) and reinforcement learning (RL) have long been studied as complementary paradigms for optimizing intelligent behaviors. Evolutionary methods excel at population-level exploration over discrete, high-level structures \citep{koza1994genetic, evolution1997handbook, lehman2011abandoning}, while reinforcement learning is well suited for fine-grained optimization of policy parameters through trial and error \citep{williams1992simple, sutton1999policy, alphago}. Historically, however, these two approaches have largely operated on different representational levels: evolution over programs \citep{langdon2013foundations} and architectures \citep{stanley2002evolving} on the one hand, and gradient-based learning over a fixed set of parameters \citep{trpo, dqn} on the other. Bridging these representational levels in a unified and scalable way remains an open challenge, particularly in the context of large language models (LLMs) \citep{gpt3, gpt4}, where both the model weights \citep{o1-system-card} and the model contexts \citep{cot, zhou2022large} critically shape behavior.

\begin{figure}[t]
  \centering
  \vspace{-15pt}
  \includegraphics[width=1.0\textwidth]{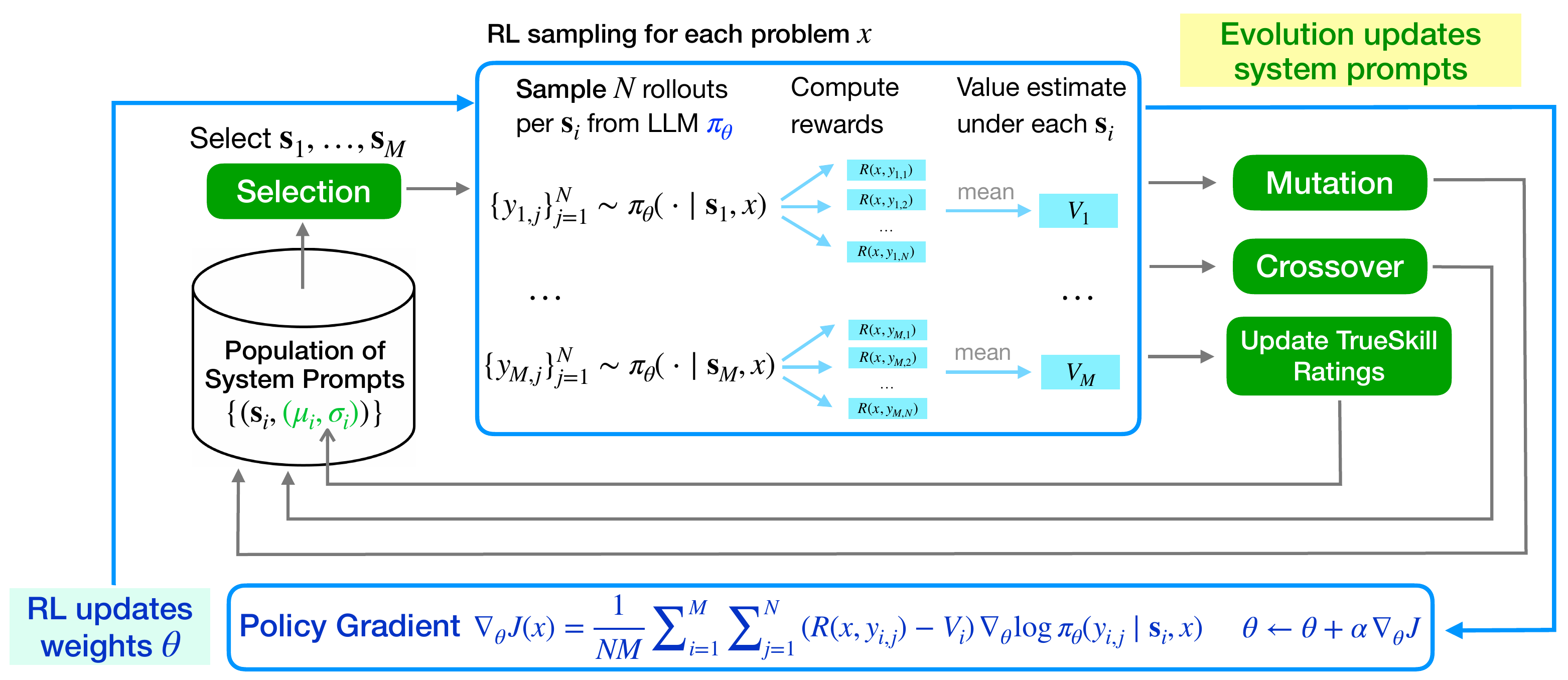}
  \vspace{-10pt}
  \caption{\textbf{Evolutionary System Prompt Learning (E-SPL)} jointly optimizes model contexts and model weights to enhance LLM self-improvement. {Evolution updates system prompts; RL updates weights. The learned system prompts can encode \textit{declarative knowledge} via articulated principles and strategies, while RL gradients can further hone the model's \textit{procedural knowledge} for reliable execution.}}
  \label{fig:method-overview}
  \vspace{-10pt}
\end{figure}

Recent work on prompt evolution and LLM agents demonstrates that prompt (or context) engineering is a powerful mechanism for improving test-time performance. Techniques such as self-reflection \citep{reflexion, self-refine, train-free-grpo} and population-based search \citep{evoprompt, promptbreeder, romera2024mathematical, gepa} show that prompts can encode substantial declarative knowledge and behavioral priors without modifying model weights. At the same time, RL has emerged as the predominant post-training approach to fundamentally improve reasoning capabilities \citep{o1-system-card, deepseek-r1, cispo, gemini-2.5}. Despite their complementary strengths, existing approaches typically optimize model contexts and weights in isolation, leaving their potential synergies unexplored.

In this work, we argue that effective self-improving agents should jointly evolve their high-level behavioral specifications and their low-level parametric policies. This perspective motivates \textbf{Evolutionary System Prompt Learning (E-SPL)} (\autoref{fig:method-overview}), a framework that tightly couples reinforcement learning over model weights with evolutionary search over system prompts, which serve as an experiential context accumulated from the model's own rollouts. By maintaining a population of system prompts that are evaluated, rated, and evolved alongside RL training, E-SPL enables structured exploration over declarative knowledge while encoding procedural knowledge into weights. This joint optimization yields synergistic dynamics in which improved prompts accelerate learning, and improved weights amplify the benefits of prompt evolution.

Experimentally, E-SPL enhances RL performance from $56.3\%\rightarrow 60.6\%$ on AIME 2025, $50.0\%\rightarrow 52.7\%$ on the HMMT 2025 November test, and notably, $38.8\% \rightarrow 45.1\%$ on AIME $\rightarrow$ BeyondAIME in an easy-to-hard generalization setting, using a base model of DeepSeek-v3.1 \citep{deepseekv3}. E-SPL significantly outperforms prompt evolution as well, showing that weight updates are often necessary for improving the model's tacit knowledge and intuition. On an agentic search task \citep{search-r1}, we improved RL performance from $44.2\% \rightarrow 48.6\%$ on OpenAI's \mbox{gpt-oss-120b} \citep{openai-gpt-oss} with limited data. These results show that combining RL and system prompt evolution unlocks a synergistic form of self-improvement that neither approach achieves in isolation.

\begin{figure}[t]
  \centering
  \vspace{-10pt}
  \includegraphics[width=0.95\textwidth]{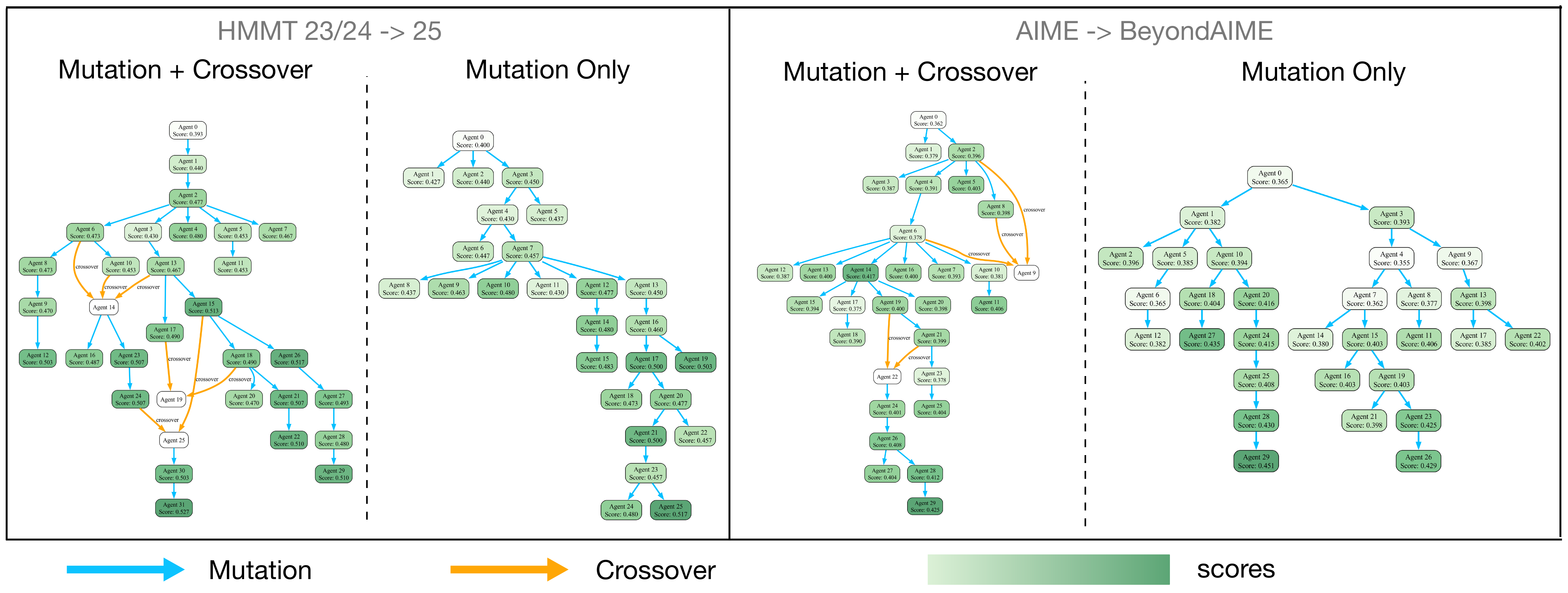}
  \vspace{-5pt}
  \caption{\textbf{Evolutionary trees of E-SPL}. During RL, E-SPL creates an evolutionary tree of system prompts, by re-using the same data already generated by RL with modest additional computational overhead. Each genetic operator (mutation or crossover) only requires a sampling server for LLM self-reflection with different context construction strategies, which can be concurrent with RL gradient updates.}
  \label{fig:evo-tree-2}
  \vspace{-2pt}
\end{figure}

\begin{figure}[!tb]
\centering
\vspace{4pt}

\begin{runcard}[title={Discovered System Prompts for Solving Math Problems (DeepSeek-v3.1)}, fontupper=\small]

{\textbf{[1]}} For generating functions with a numerator $(1 - x^M)^k$ and a denominator as a product of geometric series, the coefficient of $x^N$ is a sum over $j$ of binomial coefficients times the number of solutions to a Diophantine equation. If $N < M$, only the $j=0$ term contributes. For $N \geq M$, multiple $j$ may contribute, and the signs from the binomial coefficients must be considered.

\smallskip
\textbf{[2]} After computations, {\color{Brown}verify results for consistency across domains: check denominators, symmetry, and physical plausibility. Use alternative methods to recompute critical quantities and detect patterns such as periodicity.} 

\smallskip
\textbf{[3]} Avoid detours and unjustified guesses, including contest heuristics without validation. {\color{Brown}Trust sound derivations; if inconsistencies persist, explore alternative configurations or identities.}

\smallskip
\textbf{[4]} To verify modular injectivity, factor differences and check non-vanishing of the cofactor. For complex cases, test small primes; derivative condition is sufficient but not necessary.

\smallskip
\textbf{{[5]}} For sums over integer pairs with conditions like ${\gcd(a,b)=1}$, consider using Möbius inversion or changing the order of summation to simplify the expression.

\smallskip
\textbf{[6]} When free parameters appear but a fixed answer is required, {\color{Brown}look for cancellations forced by constraints or boundary conditions. Verify that the conclusion is consistent with all given conditions.}

\end{runcard}

\vspace{-5pt}
\caption{\centering
\textbf{Discovered strategies in learned System Prompts for solving math problems.} These explicit behavior specifications include: useful heuristics and tips for various categories of problems, self-verification strategies such as checking for consistency and plausibility, a list of common failure modes to avoid, etc. Note that RL is done under diverse system prompts, and does not overfit to any particular one.
}
\label{fig:examples-principles}
\vspace{-10pt}
\end{figure}

\begin{figure}[t]
  \centering
  \vspace{-2pt}
  \includegraphics[width=1.0\textwidth]{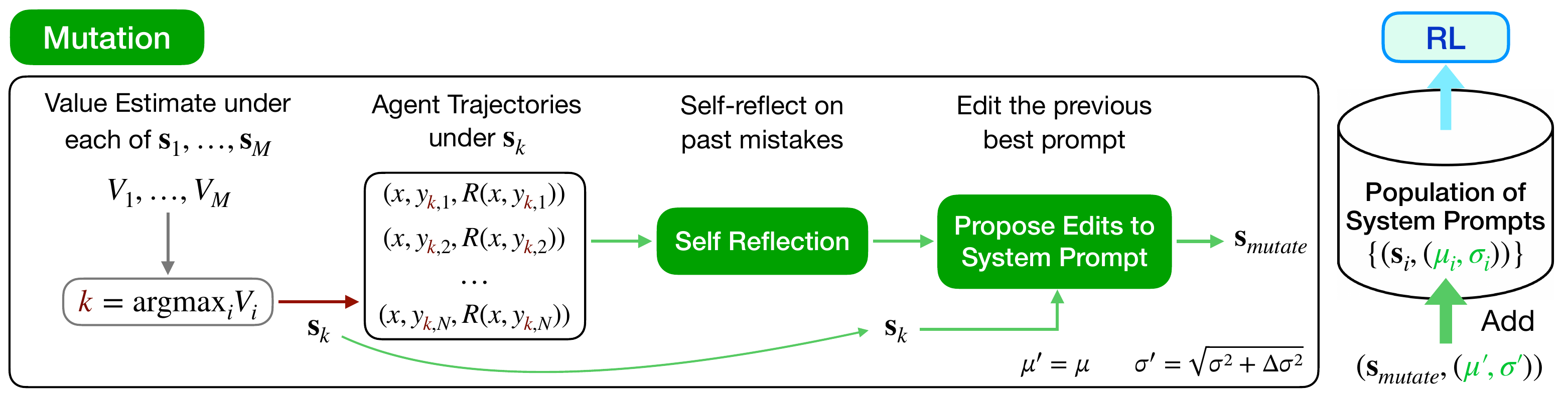}
  \vspace{-14pt}
  \caption{\textbf{Mutation operator} in E-SPL. The highest-performing prompt in each iteration undergoes \textbf{LLM self-reflection} on group-wise agent trajectories and their outcomes. An LLM-generated $\mathrm{diff}$ edits the parent into a child system prompt, removing ineffective rules and converting observed mistakes into improved declarative instructions, yielding a new prompt that enters the evolutionary population.}
  \label{fig:mutation-operator}
  \vspace{-10pt}
\end{figure}

\begin{figure}[t]
  \centering
  \vspace{2pt}
  \includegraphics[width=1.0\textwidth]{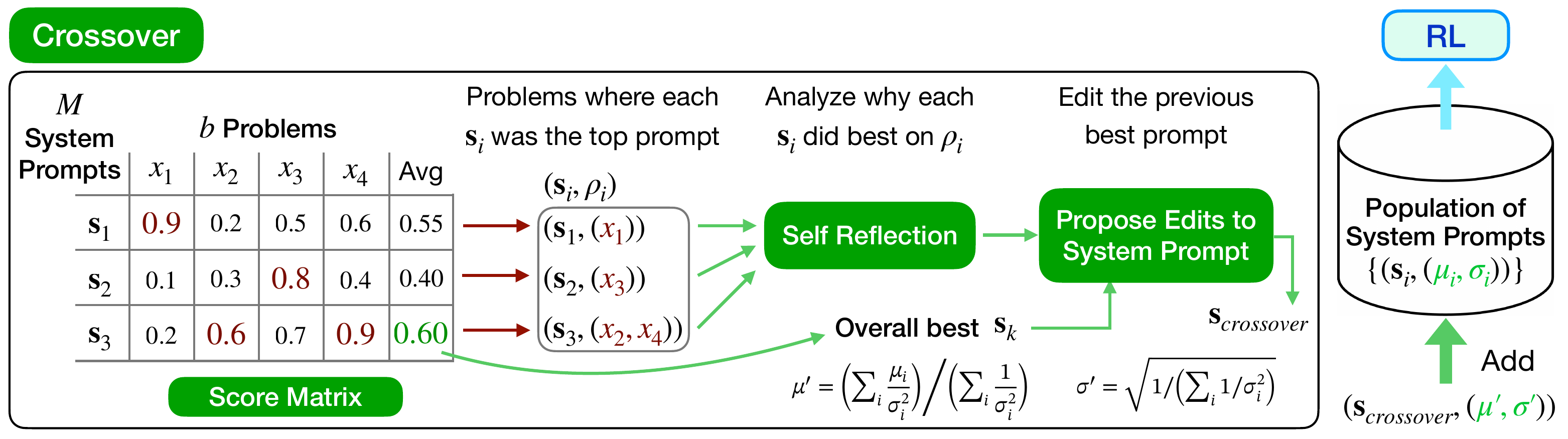}
  \vspace{-14pt}
  \caption{\textbf{Crossover operator} in E-SPL. System prompts are compared based on their problem-wise performance within the current RL batch. Guided by these differential strengths and weaknesses, an \textbf{LLM self-reflection} process selectively recombines the most effective complementary segments from multiple parent prompts, yielding a new child prompt that enters the evolutionary population.}
  \label{fig:crossover-operator}
  \vspace{-15pt}
\end{figure}

\section{Related Work}

\paragraph{Prompt Evolution and Self-Reflection}
Evolutionary and search-based approaches such as EvoPrompt \citep{evoprompt}, Promptbreeder \citep{promptbreeder}, Automatic Prompt Engineer \citep{zhou2022large}, and FunSearch \citep{romera2024mathematical} demonstrate that population-based search over prompts and programs can substantially improve test-time performance. Iterative self-reflection methods, including Reflexion \citep{reflexion}, Self-Refine \citep{self-refine}, and Training-Free GRPO \citep{train-free-grpo}, further show that LLMs can self-improve through context refinement alone. However, these approaches operate with frozen model parameters, limiting their ability to improve implicit knowledge acquired through experience.

\vspace{-5pt}
\paragraph{Scaling RL for LLMs} RL has become the primary mechanism for improving LLM reasoning, from early preference-based alignment \citep{learning-to-summarize, instructgpt} to reasoning-focused methods such as o1 \citep{o1-system-card}, R1 \citep{grpo, deepseek-r1}, and further improvements to policy gradients \citep{dapo, cispo, williams1992simple, ppo}. However, these methods assume a fixed system prompt during training, treating contexts as exogenous. While \citet{reward-guided-prompt-evolving} discover prompts within RL loops, those serve as synthetic training data rather than system-level declarative knowledge. Recent observations \citep{tan2025inoculation, azarbal2025recontextualization, macdiarmid2025natural} suggest that system-level contexts can improve generalization by reducing optimization pressure to globally update model weights.

\vspace{-5pt}
\paragraph{Evolutionary Algorithms and Genetic Programming} Hybrid Evolution+RL approaches \citep{erl, jaderberg2017population, cem-rl} leverage population dynamics to complement gradient-based optimization. More recently, FunSearch \citep{romera2024mathematical} and AlphaEvolve \citep{alphaevolve} demonstrate that LLM-guided evolutionary search over programs can yield novel discoveries. The Darwin Gödel Machine \citep{darwin-godel-machine} applies self-referential modification to agentic prompts but does not update model weights. Classical genetic programming \citep{koza1994genetic, langdon2013foundations} refines candidate programs via mutation and crossover; an LLM under a system prompt can be viewed as such a program \citep{gepa, alphaevolve}, with LLM-based genetic operators \citep{hemberg2024evolving}. Unlike prior work \citep{algo-discovery-with-llm} that combines LLM evolution with DPO \citep{dpo} for combinatorial tasks, E-SPL tackles the more general setting of LLM post-training, jointly optimizing system prompts and model weights in a single loop.

\section{Method}

E-SPL conditions RL weight updates on a population of system prompts, and reuses the resulting RL data to jointly evolve this population via two genetic operators, mutation and crossover, both of which are based on LLM self-reflection (\cref{alg:e-spl}).
\textbf{We design the evolutionary algorithm to be directly compatible with RL post-training pipelines, allowing RL to benefit from an evolving repertoire of declarative knowledge} while incurring modest additional overhead.

\vspace{-5pt}
\subsection{Reinforcement Learning for Weight Update}
The RL component of E-SPL optimizes the LLM policy $\pi_{\theta}$ to maximize rewards on dataset $\mathcal{D}$ under a variety of system prompts. Each system prompt, denoted as $\mathbf{s}$, is a list of strategic instructions that contain declarative knowledge on how to solve various categories of problems. The system prompts are not given \textit{a priori}, but are instead discovered by the LLM through self-reflection on its own experience during the RL training process. Each RL iteration uses $M$ system prompts $\{\mathbf{s}_{i}\}_{i=1}^{M}$ in parallel, so that the model weights do not overfit to any particular system prompt; this also induces direct evolutionary competition among system prompts.
For each problem $x\sim \mathcal{D}$ and each system prompt $\mathbf{s}_{i}$, we sample $N$ trajectories $y_{i,j}\sim \pi_{\theta}(\cdot \mid \mathbf{s}_{i}, x)$ for $j=1,\dots,N$, compute outcome rewards $r_{i,j} = R(x,y_{i,j})$, and value estimates $V_{i} = \frac{1}{N} \sum_{j=1}^{N} r_{i,j}$\label{eq:value-function-def}.
The policy gradient objective for each problem $x$ is $\mathcal{J}(\theta, x) = {\textstyle\frac{1}{NM}} {\textstyle\sum_{i=1}^{M}} {\textstyle\sum}_{j=1}^{N} ( r_{i,j} - V_{i} ) \log \pi_{\theta}( y_{i,j}| \mathbf{s}_{i}, x)$.
We update $\theta$ on a batch $\mathcal{B} \sim \mathcal{D}$, regularized toward $\pi_{\text{ref}}$: $\theta \leftarrow \theta + \alpha \, {\textstyle\frac{1}{ |\mathcal{B}| }} {\sum}_{x \in \mathcal{B}} \nabla_{\theta} (\mathcal{J}(\theta, x) - \beta \, \mathrm{KL}( \pi_{\theta}, \pi_{\text{ref}}))$.
When using low-rank adaptation \citep{lora}, the KL term is implicit and often ignored ($\beta=0$).

\vspace{-5pt}
\subsection{Evolutionary Algorithm for Context Update}
Integrating evolutionary search with online RL poses two challenges: prior genetic operators assume a stationary fitness landscape, whereas RL shifts model weights every iteration; and they require separate evaluation on held-out data, whereas RL generates rollouts on-policy. E-SPL addresses both by reusing RL rollouts for fitness evaluation and employing TrueSkill ratings to aggregate relative rankings across non-stationary iterations.

Our evolutionary algorithm (\textbf{EA}) optimizes the LLM system prompt to best elicit model capabilities and encode explicit knowledge learned from experience. Like in classical EA, we keep a population of candidates, and: (1) \textbf{evaluate the fitness} of different system prompts continuously during training; (2) \textbf{select} the successful ones from the population; (3) \textbf{reproduce} new system prompts from selected parents to add to the population via mutation and crossover, both grounded in LLM self-reflection but differing in their context construction strategies. The population $\mathcal{S}$ starts with a single {root}, and gradually grows into an evolutionary tree (\autoref{fig:evo-tree-2}).

\begin{wrapfigure}{r}{0.59\textwidth}
  \vspace{-18pt}
  \begin{minipage}{0.59\textwidth}
  \begin{algorithm}[H]
    \caption{Evolutionary System Prompt Learning for RL}
    \label{alg:e-spl}
    \begin{algorithmic}
      \STATE {\bfseries Input:} Problem set $\mathcal{D}(\cdot)$, reward function $R(\cdot, \cdot)$.
      \STATE {\bfseries Input:} RL policy $\pi_{\theta}(\cdot)$; reference policy $\pi_{\text{ref}}(\cdot)$.
      \STATE {\bfseries Input:} RL hyper-parameters: $N$, $\alpha$, $\beta$.
      \STATE {\bfseries Input:} Initial system prompt $\mathbf{s}_{\text{root}}$; $M$, $p_{\text{crossover}}$.
      \STATE {\bfseries Input:} Initial $(\mu_{0}, \sigma_{0})$ for {$\mathrm{TrueSkill}$}; $\Delta\sigma$; $\lambda$.
      \STATE Initialize $\mathcal{S} = \{(\mathbf{s}_{\text{root}}, \mu_{0}, \sigma_{0})\}$. \hfill \COMMENT{\textbf{Population to Evolve}}
      \WHILE{not converged}
      \STATE Sample a problem $x\sim \mathcal{D}$
      \STATE Sample $M$ system prompts $\{(\mathbf{s}_{i}, \mu_{i}, \sigma_{i})\}_{i=1}^{M} \sim \mathcal{S}$, each according to probability $p_{i} \propto \exp( \mu_{i} + \lambda \sigma_{i} )$
      \FOR{$i=1$ {\bfseries to} $M$}
          \STATE Sample $N$ solutions $\{y_{i,j}\}_{j=1}^{N}$: $y_{i, j}\sim \pi_{\theta}(\cdot \mid \mathbf{s}_{i}, x)$
          \STATE Reward $r_{i,j} = R(x, y_{i, j})$, value {$V_{i} = \frac{1}{N} {\sum}_{j=1}^{N} r_{i,j}$}
      \ENDFOR
      \STATE {$\mathcal{J}(\theta) = {\frac{1}{N M}} {\sum}_{i} {\sum}_{j} (r_{i,j} - V_{i}) \log \pi_{\theta}(y_{i,j} | \mathbf{s}_{i}, x) $}
      \STATE {$\theta \leftarrow \theta + \alpha \nabla_{\theta} (\mathcal{J}(\theta) - \beta \mathrm{KL}( \pi_{\theta}, \pi_{\text{ref}} ) )$} \hfill \COMMENT{{{\textbf{RL update}}}}
      \vspace{5pt}

      \STATE $\mathbf{v} \leftarrow \mathrm{argsort}([V_{1},\dots, V_{M}])$ \hfill \COMMENT{{\color{Brown}Rank the list of $\mathbf{s}_{i}$}}
      \STATE $\{(\mu_{i}, \sigma_{i})\}_{i=1}^{M} \leftarrow {\mathrm{TrueSkill}.\mathrm{update}}(\{(\mu_{i}, \sigma_{i})\}_{i=1}^{M}, \mathbf{v})$ \hfill \COMMENT{{\color{Brown}Change the ratings in the population $\mathcal{S}$}}
      \vspace{5pt}

      \STATE $k\leftarrow \mathrm{argmax}_{i} V_{i}$ \hfill \COMMENT{{\color{Brown}{Only {mutate} the best prompt} $\mathbf{s}_{k}$}}
      \STATE Create {\color{magenta}\textbf{{\text{mutation}} self-reflection}} prompt $\Psi_{\text{reflect}}$ using $[(x, y_{k,1}, r_{k,1}), \dots, (x, y_{k,N}, r_{k,N})]$.
      \STATE $\ell \sim \pi_{\text{ref}}(\cdot \mid \Psi_{\text{reflect}}) \quad \mathrm{diff} \sim \pi_{\text{ref}}(\cdot \mid \mathbf{s}_{k}, \ell)$
      \STATE $\mathbf{s}_{\text{mutate}} \leftarrow \mathrm{git}.\mathrm{apply}(\mathbf{s}_{k}, \mathrm{diff})$ \hfill \COMMENT{{\textbf{Mutation step}}}
      \STATE $\mathcal{S}.\mathrm{append}((\mathbf{s}_{\text{mutate}}, \mu_{k}, \sqrt{\sigma_{k}^2 + \Delta\sigma^2}))$
      \vspace{5pt}

      \IF{$p_{\text{crossover}} > \mathrm{random.rand()}$}
          \STATE Let $\mathcal{B} = \{ x_{b} \}_{b=1}^{B}$ be a batch of $B$ past problems
          \STATE Let $\Omega=\{(\mathbf{s}_{i}, \mu_i, \sigma_i)\}_{i=1}^{M}$
          \STATE $(\mathbf{s}', \mu', \sigma')=\mathrm{Crossover}(\mathcal{B}, R, \Omega, \pi_{\theta}, \pi_{\text{ref}} )$ \hfill  \COMMENT{\textbf{Alg \ref{alg:crossover}}}
          \STATE $\mathcal{S}.\mathrm{append}((\mathbf{s}', \mu', \sigma'))$
      \ENDIF
      \ENDWHILE
    \end{algorithmic}
  \end{algorithm}
  \end{minipage}
  \vspace{-25pt}
\end{wrapfigure}

\begin{figure}[t]
  \centering
  \includegraphics[width=0.92\textwidth]{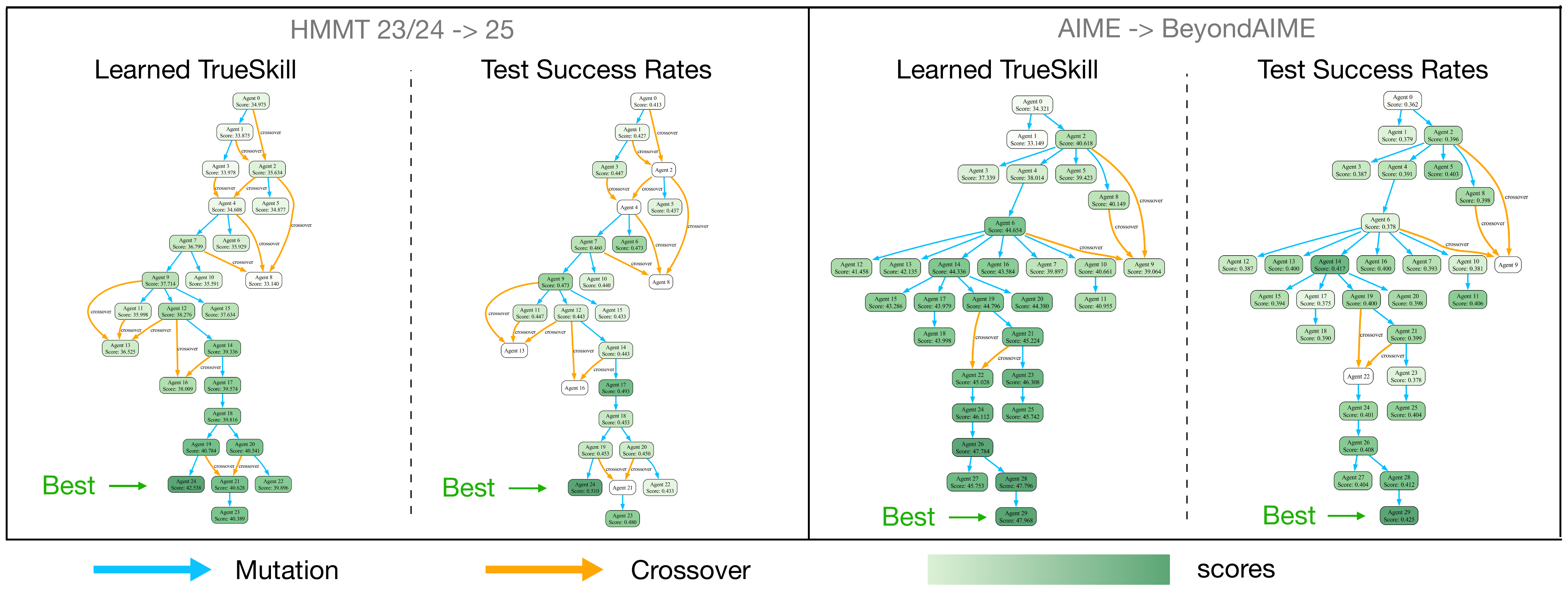}
  \vspace{-2pt}
  \caption{\textbf{TrueSkill scores learned from train-time rankings are predictive of relative test-time performance of system prompts.} E-SPL treats each RL iteration as a set of relative comparisons between system prompts for TrueSkill message passing. The visualization shows that E-SPL effectively aggregates noisy train-time comparisons into a stable and informative criterion for evolutionary selection.}
  \label{fig:evo-tree-1}
  \vspace{-10pt}
\end{figure}

\vspace{-5pt}
\paragraph{Fitness Evaluation}
Our EA reuses RL rollout statistics directly, eliminating the held-out evaluation typical of prompt evolution methods. Since absolute returns are incomparable across iterations (different batches and weights), we treat each iteration as a \textbf{tournament}: within an iteration all prompts share the same batch and weights, so only their relative ordering matters. We rank prompts by $V_i$, compute $\mathbf{v} \leftarrow \mathrm{argsort}([V_{1},\dots, V_{M}])$, and update persistent TrueSkill \citep{trueskill} ratings $(\mu_i, \sigma_i)$, i.e., each prompt's estimated skill and its uncertainty, via Bayesian message passing (see \S\ref{sec:appendix-trueskill} and \autoref{fig:evo-tree-1}).

\vspace{-5pt}
\paragraph{Selection} We sample $M$ prompts per iteration using an optimistic UCB-style score: $\mathrm{score}_{i} = \mu_{i}+\lambda \sigma_{i}$, $p_{i} \propto \exp ( \mathrm{score}_{i} /T )$\label{eq:selection-equation}, where $\lambda$ and $T$ trade off exploitation and exploration. Prompts with high mean performance are exploited, while those with high uncertainty are explored.

\vspace{-5pt}
\paragraph{Mutation} We mutate only the best-performing prompt $\mathbf{s}_k$ ($k \leftarrow \mathrm{argmax}_{i} V_{i}$; \autoref{fig:mutation-operator}), creating direct evolutionary pressure: \textbf{only the prompt that outperforms all competitively selected peers reproduces}. The reference policy $\pi_{\text{ref}}$ first self-reflects on the rollouts under $\mathbf{s}_k$ to produce lessons $\ell \sim \pi_{\text{ref}}(\cdot \mid \Psi_{\text{reflect}}, \{(x, y_{k,j}, r_{k,j})\}_{j=1}^{N})$, then proposes edits $\mathrm{diff} \sim \pi_{\text{ref}}(\cdot \mid \mathbf{s}_{k}, \ell)$ applied as $\mathbf{s}_{\text{mutate}} \leftarrow \mathrm{git}.\mathrm{apply}(\mathbf{s}_{k}, \mathrm{diff})$. The child is added to $\mathcal{S}$ with rating $(\mu_{k}, \sqrt{\sigma_{k}^2+\Delta\sigma^2})$; inheriting the parent's rating with increased uncertainty favors exploration.

\vspace{-5pt}
\paragraph{Crossover} While mutation evolves individual lineages, crossover recombines knowledge across parents (\autoref{fig:crossover-operator}). We evaluate $M$ prompts on a batch $\mathcal{B}$ of $B$ problems, yielding a score matrix $\Phi \in \mathbb{R}^{M\times B}$ where $\Phi_{i,b}$ captures the per-problem performance of each prompt. The problem-wise strengths identified by $\Phi$ are incorporated into the crossover reflection prompt $\Psi_{\mathrm{cr}}$, grounding self-reflection in concrete differential advantages and enabling targeted recombination. The resulting $\mathbf{s}_{\text{crossover}}$ is added to $\mathcal{S}$ with a precision-weighted average of the parents’ ratings (\cref{alg:crossover}).

\vspace{-5pt}
\paragraph{Summary} E-SPL augments RL post-training with an evolutionary algorithm based on self-reflection from experience. In parallel with RL updates, E-SPL uses relative performance rankings of system prompts to inform evolutionary selection. Mutation and crossover operators, both driven by LLM self-reflection, produce new system prompts with modest computational overhead.

\FloatBarrier
\section{Experiments}

\begin{figure}[!t]
  \centering
  \vspace{0pt}
  \includegraphics[width=1.0\textwidth]{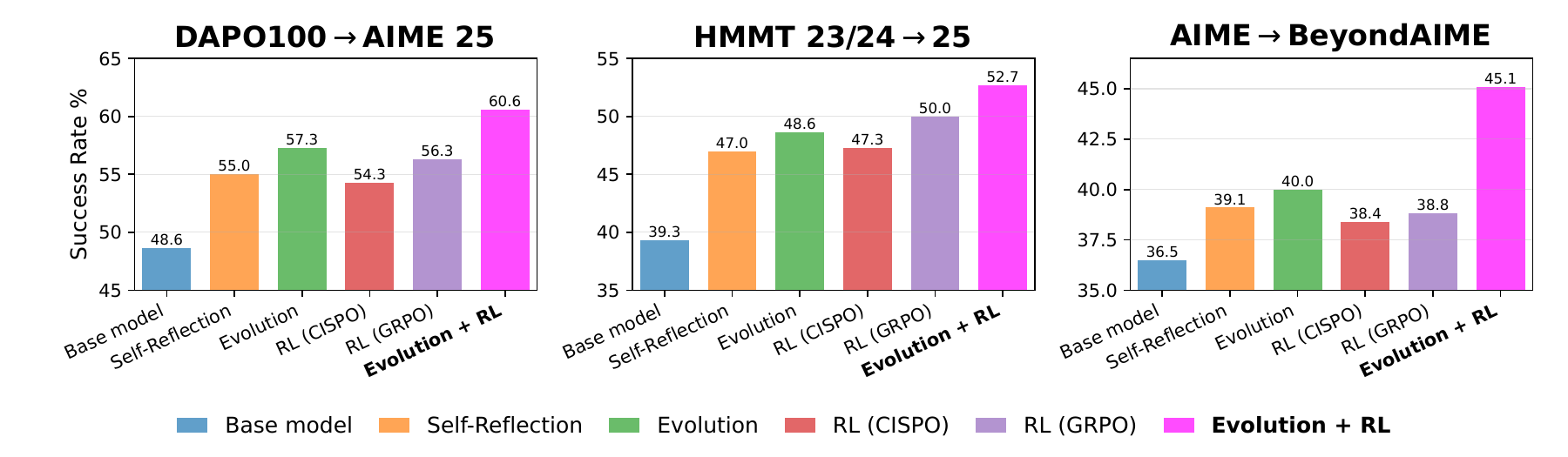}
  \vspace{-12pt}
  \caption{\centering
  Results on mathematical reasoning datasets using DeepSeek-v3.1. \textbf{Evolution + RL} (our E-SPL method) performs best compared to self-reflection (prompting only), evolution (prompting only), and RL (weight update only). The results show that coupling RL with evolution on system prompts unlocks a synergistic form of self-improvement that neither approach achieves in isolation.}
  \label{fig:main-results}
  \vspace{-15pt}
\end{figure}

We evaluate E-SPL on mathematical reasoning and agentic search benchmarks, asking: \textbf{(1)} Does jointly evolving system prompts and model weights outperform optimizing either alone? \textbf{(2)} How does E-SPL affect the learning speed and generalization of RL? \textbf{(3)} What emergent patterns arise in discovered system prompts?

\vspace{-5pt}
\subsection{Experimental Setup}

\paragraph{Tasks and Datasets}
Our main evaluations include: \textbf{(1) DAPO100$\rightarrow$AIME25}, which follows the protocols from \citep{train-free-grpo} that trains on $100$ selected problems from the DAPO dataset \citep{dapo} and tests on the AIME 2025 competition; \textbf{(2) HMMT 23/24$\rightarrow$25}, which uses all of HMMT 2023, 2024, and February 2025 problems as the training set, and November 2025 problems as the test set; \textbf{(3) AIME$\rightarrow$BeyondAIME}, which uses past years of AIME (AIME 22, AIME 23, and AIME 24) as the training set, and BeyondAIME \citep{beyondaime} as the test set, a set of problems at or above AIME difficulty. In addition, we also evaluate on \textbf{(4) Agentic Search} on Natural Questions (NQ) \citep{nq} and HotpotQA \citep{hotpotqa}, where we subsample 196 problems for training and 64 problems for testing; the task is multi-hop question answering with search engine API calls within the chain-of-thoughts \citep{search-r1}.

\vspace{-6pt}
\paragraph{Models} On math tasks, we use DeepSeek-v3.1 \citep{deepseekv3}, a 671B MoE \citep{moe} model with 37B active parameters. On agentic search tasks, we use OpenAI's gpt-oss-120b \citep{openai-gpt-oss}, which is a 120B MoE model with 5B active parameters. All finetuning is done via low-rank adaptation \citep{lora} with a rank of $32$.

\vspace{-6pt}
\paragraph{Methods Compared} We compare the following variants:
\vspace{-6pt}
\begin{itemize}
    \item \textbf{RL only}: standard RL post-training with a fixed system prompt; we benchmark the popular \textbf{GRPO} \citep{grpo} and \textbf{CISPO} \citep{cispo} algorithms.
    \item \textbf{Self-reflection}: iterative prompt refinement via reflection \citep{reflexion, train-free-grpo, reasoningbank}, without weight updates. Equivalent to mutation-only with frozen weights and population size 1.
    \item \textbf{Evolution}: our evolutionary algorithm (EA) for system prompt search with frozen model weights. Our EA is similar to GEPA \citep{gepa} but uses only train-time rollout statistics rather than a separate feedback set.
    \item \textbf{Evolution + RL (E-SPL)}: our full method, jointly evolving contexts and updating weights.
\end{itemize}

\begin{figure}[t]
  \centering
  \includegraphics[width=0.98\textwidth]{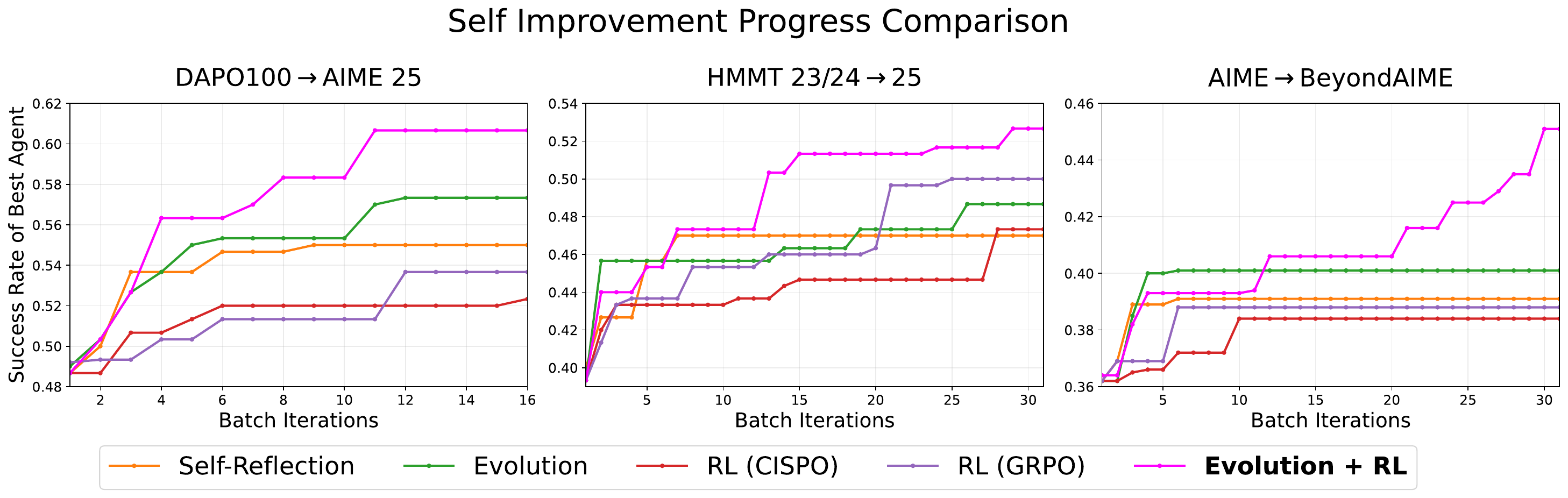}
  \vspace{2pt}
  \caption{\textbf{E-SPL learns faster, and achieves better asymptotic performance that neither RL nor prompt evolution alone could achieve.} By studying the learned system prompts (e.g., \autoref{fig:examples-principles} and \autoref{fig:example-e-spl-1}), we hypothesize that E-SPL achieves this by encouraging a natural division between declarative knowledge (factual information or global heuristics that can be verbally explained) in prompts and procedural knowledge (practical, instinctive ``know-how'' that relies on intuition developed through repetition) in weights.}
  \label{fig:learning-progress}
  \vspace{-15pt}
\end{figure}

\vspace{-8pt}
\subsection{Main Results}
\label{sec:main-results}

\begin{wrapfigure}{r}{0.4\textwidth}
  \vspace{-16pt}
  \centering
  \includegraphics[width=0.38\textwidth]{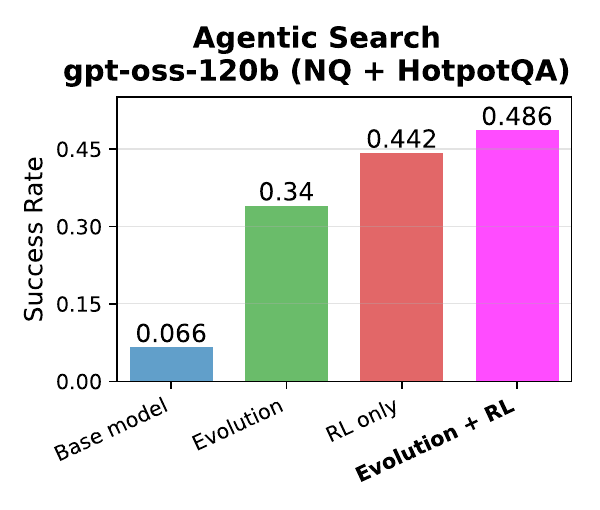}
  \vspace{-8pt}
  \caption{{E-SPL outperforms both Evolution and RL on agentic search tasks with gpt-oss-120b.}}
  \label{fig:agentic-search}
  \vspace{-16pt}
\end{wrapfigure}

Figure \ref{fig:main-results} summarizes performance on math reasoning benchmarks using the DeepSeek-v3.1 model. Figure \ref{fig:agentic-search} summarizes our results on agentic search tasks using OpenAI gpt-oss-120b. Across all datasets, \textbf{Evolution + RL} (E-SPL) consistently outperforms baselines that optimize either prompts (evolution-only) or weights (RL-only) in isolation.
\vspace{-5pt}
\begin{itemize}
    \item On \textbf{AIME 2025}, E-SPL improves success rate from ${56.3\% \rightarrow 60.6\%}$ over RL-only.
    \item On \textbf{HMMT 2025}, performance improves from ${50.0\% \rightarrow 52.7\%}$ after applying E-SPL.
    \item On an easy-to-hard generalization \citep{easy-to-hard} setting, \textbf{AIME $\rightarrow$ BeyondAIME}, the gains are most pronounced, where E-SPL improves RL from ${38.8\%\rightarrow45.1\%}$ in success rate, capable of learning more generalizable strategies from easier training problems.
    \item On \textbf{agentic search} tasks, E-SPL improves RL from $44.2\% \rightarrow 48.6\%$, which demonstrates that E-SPL is not limited to mathematical reasoning, but is also effective in agentic reasoning with tool use, where evolved prompts discover novel workflows (\autoref{fig:prompt-evolution-search}).
    \item On all tasks above, E-SPL outperforms evolution-only as well, with relative improvements over evolution-only ranging from $+5.8\%$ to $+12.8\%$ on math reasoning tasks, and $+42.9\%$ on the agentic search task.
    \item E-SPL also outperforms a \textbf{compute-matched} GRPO baseline ($+4.4\%$ and $+15.1\%$ relative on HMMT and BeyondAIME; \autoref{tab:compute-matched}) and sequential evolution-then-RL training ($+4.3\%$ on AIME 2025). Tripling the GRPO group size yields diminishing returns ($+1.0\%$ on both), and E-SPL's BeyondAIME gains hold across seeds (\S\ref{sec:appendix-compute-matched}--\ref{sec:appendix-sequential}).
\end{itemize}
\vspace{-5pt}

Evolution-only consistently outperforms self-reflection, and can sometimes outperform RL-only. On both DAPO100 $\rightarrow$ AIME25 and AIME $\rightarrow$ BeyondAIME, evolution outperforms RL; on HMMT 23/24 $\rightarrow$ 25, GRPO slightly outperforms evolution. This is consistent with prior findings in GEPA \citep{gepa}.

\begin{table*}[t]
\centering
\small
\vspace{0pt}
\setlength{\tabcolsep}{3.5pt}
\renewcommand{\arraystretch}{1.05}
\begin{tabular}{%
  @{}p{0.12\textwidth}
  >{\raggedright\arraybackslash\renewcommand{\baselinestretch}{1.0}\selectfont}p{0.85\textwidth}@{}
}

\toprule
\textbf{\color{Brown}Discovered}
&
For a polynomial $P(n) = {\textstyle\prod}_{i=1}^m (n - a_i)$, the largest integer $d$ that divides $P(n)$ for all integers $n$ is $d = {\textstyle\prod}_{p} p^{e_p}$, {\color{Brown} where
$e_p = \min_{r \bmod p} \bigl(\#\{\, i : a_i \equiv r \!\!\pmod p \,\}\bigr).$
Only primes $p \le \max(a_i)$ need be considered, as for larger $p$, the $a_i$ are distinct modulo $p$, so $e_p = 0$.}
For primes where the function mapping $i$ to $a_i \bmod p$ is a permutation (e.g., if $a_i = i^3$ and $p \equiv 2 \bmod 3$), {\color{Brown}$e_p = {{\left\lfloor \frac{m}{p} \right\rfloor}}$.
} \\
\addlinespace[6pt]
\textbf{\color{MidnightBlue}Corrected} 
&
For a polynomial $P(n) = \prod_{i=1}^m (n - a_i)$, the largest integer $d$ that divides $P(n)$ for all integers $n$ is $d = \prod_{p} p^{e_p}$, {\color{MidnightBlue} where $e_p = \min_{n \in \mathbb{Z}} v_p\!\bigl(P(n)\bigr).$ Moreover, one always has the lower bound $e_p \ge \min_{r \bmod p} \bigl(\#\{\, i : a_i \equiv r \!\!\pmod p \,\}\bigr).$ In particular, for this mod-$p$ bound it suffices to consider primes $p \le m$, since if $p>m$ then some residue class occurs zero times and the minimum is $0$.} 
For primes where the function mapping $i$ to $a_i \bmod p$ is a permutation (e.g., if $a_i = i^3$ and $p \equiv 2 \bmod 3$), the bound gives {\color{MidnightBlue}$e_p \boldsymbol{\ge} \left\lfloor \frac{m}{p} \right\rfloor$.
} \\
\midrule
\textbf{\color{Brown}Discovered}
&
For graphs with odd-degree vertices, find a minimum \(T\)-join by {\color{Brown}pairing odd vertices along efficient paths}. In symmetric graphs such as grids, {\color{Brown}pair boundary vertices} optimally to minimize total length.
\\
\addlinespace[6pt]
\textbf{\color{MidnightBlue}Corrected} 
&
For graphs with odd-degree vertices, find a minimum $T$-join by {\color{MidnightBlue}pairing odd vertices via a minimum-weight perfect matching on shortest-path distances}. In symmetric graphs such as grids, {\color{MidnightBlue}consider pairings only up to symmetry} to minimize total length.
\\
\bottomrule
\end{tabular}
\vspace{0pt}
\caption{
The discovered system prompts sometimes include \textbf{heuristics that are {\color{Brown}not always true} but still helpful to problem-solving}. Unlike RL updates to model weights, the learned system prompts are interpretable and thus can be monitored and corrected. Here we showcase two subtle mistakes in two discovered principles, and what their {\color{MidnightBlue}\textbf{corrected}} versions should be (see \S\ref{app:mistake-made-by-principle} for explanations).}
\vspace{2pt}
\label{fig:corrected-principles}
\end{table*}

Figure \ref{fig:learning-progress} shows that evolution-only learns fast early on but plateaus, while E-SPL achieves both faster learning and higher asymptotic performance.

Notably, the relative strength of RL versus evolution varies by domain --- RL dominates in agentic search ($44.2\%$ vs $34.0\%$) but is on par with evolution in math reasoning. E-SPL combines the strengths of both regardless of domain.

\vspace{-5pt}
\subsection{Emergent Patterns in Learned System Prompts}

\paragraph{Generative Self-Verification} A recurring theme is generative self-verification \citep{gen-verifier}. In agentic search, the learned system prompt (\autoref{fig:prompt-evolution-search}) specifies a workflow that labels each search result as VERIFIED or NOT VERIFIED based on source reliability and verbatim phrase matching, looping until verification succeeds. In math, a similar strategy emerged: \emph{`perform sanity checks: test small instances, ensure bounds, and verify extreme cases'} (\autoref{fig:example-e-spl-2}). RL then refines execution of these workflows rather than discovering them from scratch.

\vspace{-6pt}
\paragraph{Codified Domain Expertise} E-SPL also codifies domain-specific heuristics — e.g., when to apply the Chinese Remainder Theorem or Möbius inversion (\autoref{fig:example-e-spl-1}, \autoref{fig:example-e-spl-4-short}). These declarative strategies let RL focus on honing core reasoning skills rather than memorizing domain knowledge.

\vspace{-6pt}
\paragraph{Failure Mode: Missing Caveats} Not all discovered heuristics are strictly correct; some over-generalize or miss caveats (\autoref{fig:corrected-principles}). However, unlike opaque RL weight updates, E-SPL's learned knowledge is explicit and editable, making it amenable to human monitoring, correction, and scalable oversight \citep{scalable-agent-alignment, scalable-oversight}.

\vspace{-5pt}
\subsection{Ablation Studies}

\begin{wrapfigure}{r}{0.4\textwidth}
  \vspace{-16pt}
  \centering
  \includegraphics[width=0.38\textwidth]{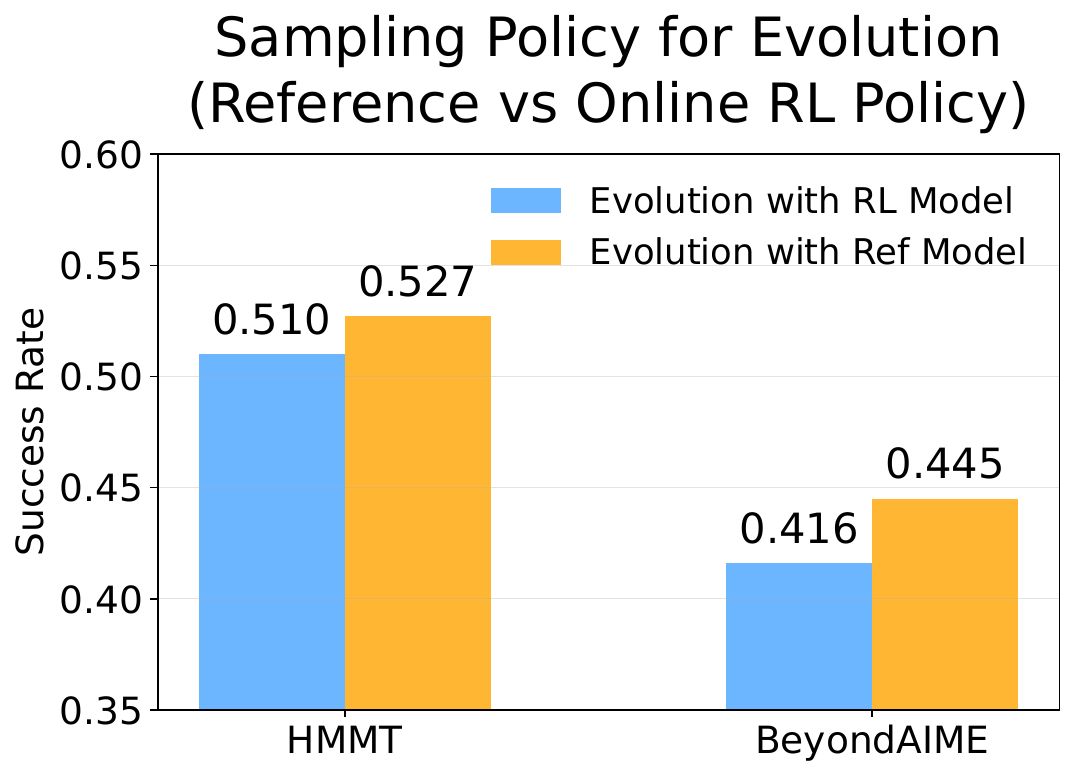}
  \caption{{Using a fixed reference policy rather than the online RL policy for prompt evolution is important.}}
  \label{fig:evo-pol-ablation}
  \vspace{-16pt}
\end{wrapfigure}

\paragraph{Sampling Policy for Evolution}
Figure \ref{fig:evo-pol-ablation} shows that using a fixed reference policy $\pi_{\text{ref}}$ for self-reflection consistently outperforms using the continually changing RL policy $\pi_{\theta}$. A fixed reference provides a stable basis for mutation and crossover, whereas allowing the editor to drift alongside the policy introduces instability.

\vspace{-5pt}
\paragraph{Effect of Crossover} Adding crossover to mutation propagates useful substructures across lineages, which may be advantageous in some domains, although final performance is comparable in our experiments (\S\ref{sec:appendix-crossover}). Both mutation-only and mutation+crossover consistently outperform RL-only, confirming that E-SPL's gains are attributable to evolutionary search over system prompts as a whole, rather than any specific operator.

\vspace{-5pt}
\paragraph{Hyperparameter Sensitivity} TrueSkill is used off-the-shelf with default parameters (see \S\ref{sec:appendix-trueskill}). The remaining hyperparameters (see \S\ref{sec:appendix-hyperparameters}) are either standard RL hyperparameters or EA hyperparameters shared identically with baselines, so E-SPL introduces no additional tuning burden beyond the baselines of RL and EA.

\begin{figure}[!tb]
\centering
\vspace{-2pt}
\caption{\textbf{An example of how the RL model utilizes learned system prompts} (also see \autoref{fig:example-e-spl-4}).}
\vspace{-8pt}
\begin{runcard}[title={Example of How E-SPL Shapes Model Behavior}, fontupper=\small]

\smallskip

\begin{runbox}[title={\textbf{Learned System Prompt}}]
When solving math problems, you MUST first carefully read and understand the helpful instructions and principles below:

$\cdots$ \quad {\color{Brown}\hl{{[G2]}}. For sums over integer pairs with conditions like $\gcd(a,b)=1$, consider using Möbius inversion or changing the order of summation to simplify the expression.} \quad $\cdots$

\end{runbox}

\begin{runbox}[title={Problem}]
Let $S$ be the set of all positive integers less than $143$ that are relatively prime to $143$. Compute the number of ordered triples $(a,b,c)$ of elements of $S$ such that $a+b=c$.

\end{runbox}

\begin{runbox}[title={Model Output}]

Let's solve the problem step by step:

\textbf{Step 1}: Factorize \( 143 = 11 \times 13 \). So, \( S \) consists of positive integers \( < 143 \) that are coprime to both $11$ and $13$.

\quad $\cdots$

\textbf{Step 5}: {{\color{Brown}Use Möbius inversion (Principle \hl{[G2]}).} Let \( f(n) \) be the number of pairs \( (a, b) \in S \times S \) with \( a + b = n \). We want \( \sum_{n=1}^{142} f(n) \cdot \mathbf{1}_{\gcd(n,143)=1} \). \quad $\cdots$}

\quad $\cdots$

\textbf{Step 11}: Apply inclusion-exclusion. \quad $\cdots$ \quad Therefore, the number of ordered triples $(a,b,c)$ is $5940$.

\textbf{Final Answer}: $\boxed{5940}$

\end{runbox}

\textbf{Ground truth answer:} $5940$
\end{runcard}
\label{fig:example-e-spl-4-short}

\vspace{-14pt}
\end{figure}

\vspace{-4pt}
\section{Conclusion}
We have shown that RL and evolutionary prompt search are deeply synergistic: E-SPL jointly optimizes model weights and system prompts, yielding consistent gains in sample efficiency, asymptotic performance, and generalization. As an experiential context, system prompts provide a natural substrate for declarative knowledge that complements procedural knowledge from RL, and the two forms of self-improvement reinforce each other. One limitation is that the self-reflection module does not yet self-improve alongside the policy, which we leave to future work. E-SPL also requires a reference model that is sufficiently capable of self-reflection and instruction following (\S\ref{sec:appendix-discussion}). Additionally, studying the safety properties of autonomous self-rewrite mechanisms such as E-SPL remains an important open direction. We hope E-SPL motivates further research on unifying evolutionary self-rewrite with gradient-based learning.

\begin{ack}
We thank the anonymous reviewers for their constructive feedback. Bridgewater AIA Labs provided resources for this work. L.Z. was supported by the Schwartz Reisman Institute for Technology and Society at the University of Toronto through a Schwartz Reisman Fellowship. The authors declare no competing interests.
\end{ack}

\bibliography{e_spl}
\bibliographystyle{plainnat}

\newpage
\section*{NeurIPS Paper Checklist}

\begin{enumerate}

\item {\bf Claims}
    \item[] Question: Do the main claims made in the abstract and introduction accurately reflect the paper's contributions and scope?
    \item[] Answer: \answerYes{}
    \item[] Justification: The abstract and introduction clearly state the contributions and report empirical improvements that are supported by the experimental results in Section~4.

\item {\bf Limitations}
    \item[] Question: Does the paper discuss the limitations of the work performed by the authors?
    \item[] Answer: \answerYes{}
    \item[] Justification: The paper discusses failure modes of learned system prompts (Section~4.3) and other limitations in the conclusion and Appendix~\ref{sec:appendix-discussion}.

\item {\bf Theory assumptions and proofs}
    \item[] Question: For each theoretical result, does the paper provide the full set of assumptions and a complete (and correct) proof?
    \item[] Answer: \answerNA{}
    \item[] Justification: The paper does not include formal theoretical results or proofs. It is primarily an empirical contribution with algorithmic descriptions.

    \item {\bf Experimental result reproducibility}
    \item[] Question: Does the paper fully disclose all the information needed to reproduce the main experimental results of the paper to the extent that it affects the main claims and/or conclusions of the paper (regardless of whether the code and data are provided or not)?
    \item[] Answer: \answerYes{}
    \item[] Justification: The paper provides full algorithmic descriptions (Algorithm~1, Algorithm~2), model specifications, dataset details, hyperparameters, and the evolutionary operator designs.

\item {\bf Open access to data and code}
    \item[] Question: Does the paper provide open access to the data and code, with sufficient instructions to faithfully reproduce the main experimental results, as described in supplemental material?
    \item[] Answer: \answerYes{}
    \item[] Justification: Code is publicly available at \url{https://github.com/LunjunZhang/E-SPL} with instructions for reproducing the main experiments. All datasets used are public.

\item {\bf Experimental setting/details}
    \item[] Question: Does the paper specify all the training and test details (e.g., data splits, hyperparameters, how they were chosen, type of optimizer) necessary to understand the results?
    \item[] Answer: \answerYes{}
    \item[] Justification: Section~4.1 specifies models, datasets, train/test splits, LoRA rank, and baselines. The appendix provides additional hyperparameters.

\item {\bf Experiment statistical significance}
    \item[] Question: Does the paper report error bars suitably and correctly defined or other appropriate information about the statistical significance of the experiments?
    \item[] Answer: \answerNo{}
    \item[] Justification: Due to the high computational cost of RL post-training, we report single-run results for most experiments; Appendix~\ref{sec:appendix-compute-matched} reports E-SPL over 3 random seeds.

\item {\bf Experiments compute resources}
    \item[] Question: For each experiment, does the paper provide sufficient information on the computer resources (type of compute workers, memory, time of execution) needed to reproduce the experiments?
    \item[] Answer: \answerYes{}
    \item[] Justification: The appendix describes the training details and computational cost (Appendix~\ref{sec:appendix-cost}). All finetuning uses LoRA with rank 32 on the specified models.

\item {\bf Code of ethics}
    \item[] Question: Does the research conducted in the paper conform, in every respect, with the NeurIPS Code of Ethics \url{https://neurips.cc/public/EthicsGuidelines}?
    \item[] Answer: \answerYes{}
    \item[] Justification: The research conforms with the NeurIPS Code of Ethics. It involves training LLMs on mathematical reasoning and question answering using publicly available datasets.

\item {\bf Broader impacts}
    \item[] Question: Does the paper discuss both potential positive societal impacts and negative societal impacts of the work performed?
    \item[] Answer: \answerYes{}
    \item[] Justification: The paper discusses the interpretability advantages of E-SPL (Section~4.3). Improved LLM reasoning carries the dual-use considerations common to all LLM research.

\item {\bf Safeguards}
    \item[] Question: Does the paper describe safeguards that have been put in place for responsible release of data or models that have a high risk for misuse (e.g., pre-trained language models, image generators, or scraped datasets)?
    \item[] Answer: \answerNA{}
    \item[] Justification: The paper proposes a training method rather than releasing a pre-trained model. The base models used (DeepSeek-v3.1, gpt-oss-120b) are already publicly available.

\item {\bf Licenses for existing assets}
    \item[] Question: Are the creators or original owners of assets (e.g., code, data, models), used in the paper, properly credited and are the license and terms of use explicitly mentioned and properly respected?
    \item[] Answer: \answerYes{}
    \item[] Justification: All models and datasets used are publicly available, credited, and used in accordance with their terms of use.

\item {\bf New assets}
    \item[] Question: Are new assets introduced in the paper well documented and is the documentation provided alongside the assets?
    \item[] Answer: \answerYes{}
    \item[] Justification: Code is publicly available at \url{https://github.com/LunjunZhang/E-SPL} with documentation for reproducing experiments.

\item {\bf Crowdsourcing and research with human subjects}
    \item[] Question: For crowdsourcing experiments and research with human subjects, does the paper include the full text of instructions given to participants and screenshots, if applicable, as well as details about compensation (if any)?
    \item[] Answer: \answerNA{}
    \item[] Justification: The paper does not involve crowdsourcing or research with human subjects.

\item {\bf Institutional review board (IRB) approvals or equivalent for research with human subjects}
    \item[] Question: Does the paper describe potential risks incurred by study participants, whether such risks were disclosed to the subjects, and whether Institutional Review Board (IRB) approvals (or an equivalent approval/review based on the requirements of your country or institution) were obtained?
    \item[] Answer: \answerNA{}
    \item[] Justification: The paper does not involve research with human subjects.

\item {\bf Declaration of LLM usage}
    \item[] Question: Does the paper describe the usage of LLMs if it is an important, original, or non-standard component of the core methods in this research? Note that if the LLM is used only for writing, editing, or formatting purposes and does \emph{not} impact the core methodology, scientific rigor, or originality of the research, declaration is not required.
    \item[] Answer: \answerYes{}
    \item[] Justification: LLMs are a core component of E-SPL: the method uses LLMs both as the policy being trained and as the reference model for self-reflection (Section~3).

\end{enumerate}

\newpage
\appendix
\section{Method Details}

\subsection{Overview of the Genetic Operators}
\label{sec:appendix-self-reflection}

In every RL iteration, E-SPL turns the rollouts it has just generated into new system prompts. The naive approach of handing all rollouts to the reference model and asking it to ``improve this prompt'' faces three difficulties, which shape the design of the genetic operators:
\begin{enumerate}
    \item \emph{Noisy evidence.} An iteration produces many long reasoning traces with only outcome rewards, so which instructions helped or hurt is never directly observed.
    \item \emph{A shared prompt.} The system prompt is shared across all problems, and the policy is trained under it: an edit that fixes one problem can hurt others, and a wholesale rewrite can discard principles accumulated over earlier generations.
    \item \emph{Conflicting lessons.} Lessons drawn from different problems can overlap or contradict each other, and keeping all of them would make the prompt grow without bound.
\end{enumerate}

\begin{figure}[t]
  \centering
  \vspace{0pt}
  \includegraphics[width=0.91\textwidth]{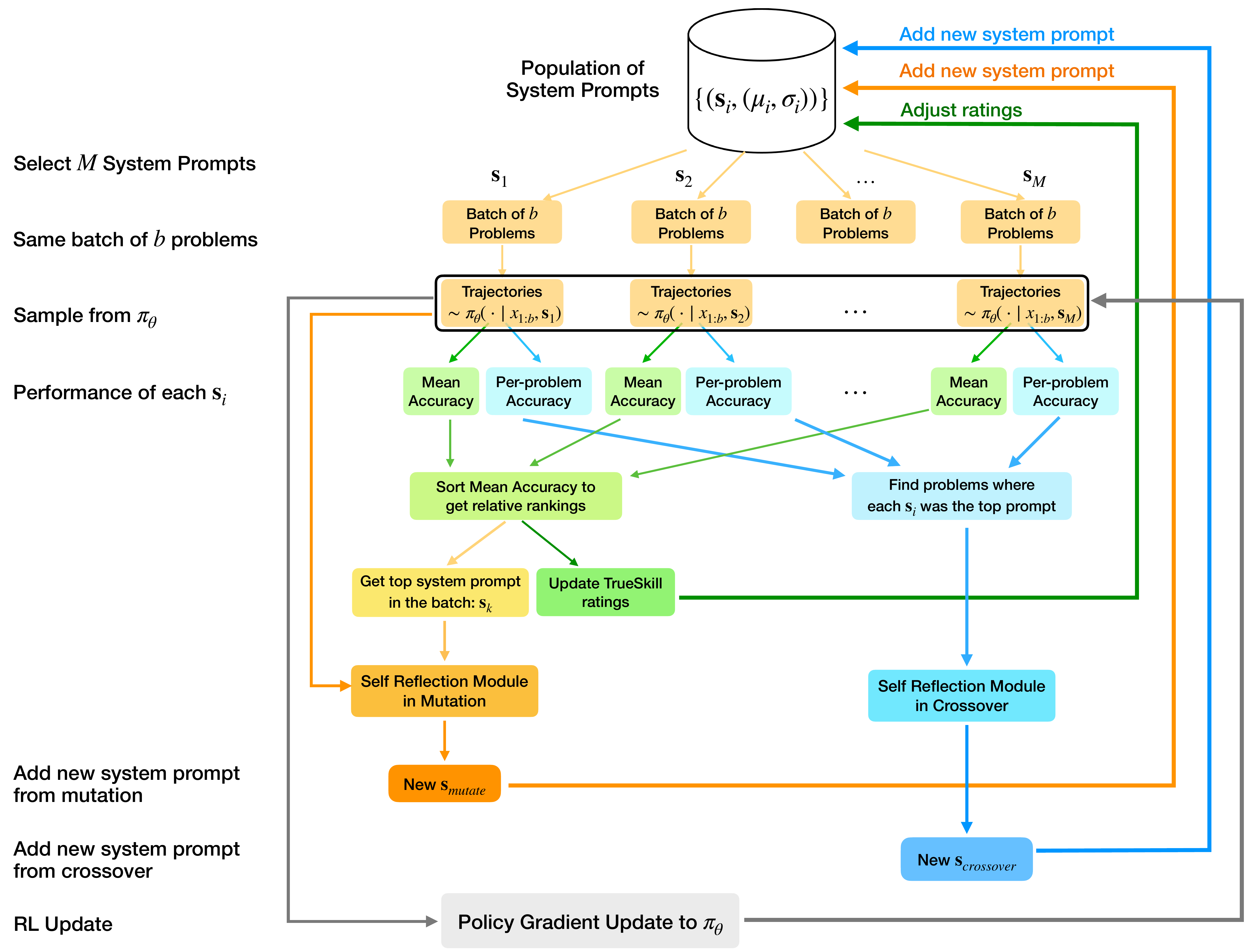}
  \vspace{10pt}
  \caption{\textbf{Detailed overview of the E-SPL pipeline}, which illustrates how \autoref{fig:method-overview}, \autoref{fig:mutation-operator}, and \autoref{fig:crossover-operator} fit into a single algorithm.}
  \label{fig:e-spl-pipeline-details}
\end{figure}

\begin{figure}[t]
  \centering
  \includegraphics[width=1.0\textwidth]{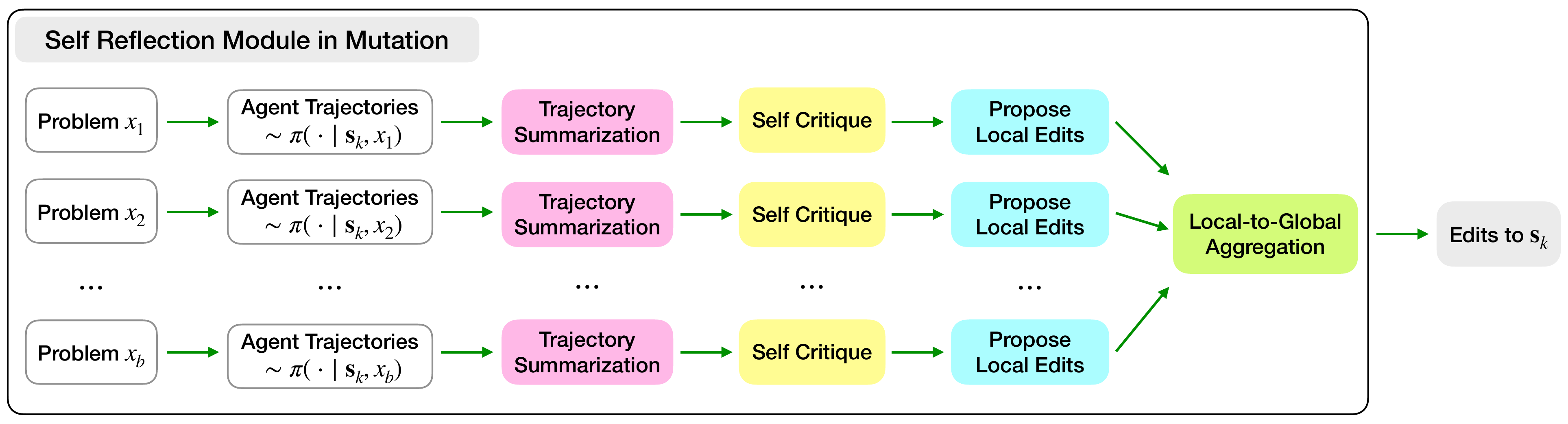}
  \vspace{-10pt}
  \caption{\textbf{Self-Reflection Module in Mutation} (where the best-performing system prompt $\mathbf{s}_{k}$ is edited) consists of four sequential modules: trajectory summarization (\S\ref{sec:appendix-trajectory-summary}), self-critique (\S\ref{sec:appendix-local-critique}), local edit proposal (\S\ref{sec:appendix-max-ops}), and local-to-global aggregation (\S\ref{sec:appendix-global-revision}).}
  \label{fig:self-reflection-operator}
  \vspace{0pt}
\end{figure}

E-SPL implements both mutation and crossover as LLM-driven editing operators over past system prompts; \autoref{fig:e-spl-pipeline-details} shows how both operators fit into the overall pipeline. The mutation operator consists of (i) a self-reflection stage that diagnoses performance from RL rollouts, which has two steps: trajectory summarization (\S\ref{sec:appendix-trajectory-summary}) and self-critique (\S\ref{sec:appendix-local-critique}); followed by (ii) a patch synthesis stage that applies structured edits to the parent prompt using a git-style diff, which has two steps: local edit proposal (\S\ref{sec:appendix-max-ops}) and local-to-global aggregation (\S\ref{sec:appendix-global-revision}). \autoref{fig:self-reflection-operator} illustrates these four steps. On the other hand, the crossover operator has a single stage (\S\ref{sec:appendix-crossover-operator}, \cref{alg:crossover}).

Each step of mutation addresses one of these difficulties:
\begin{itemize}
    \item Trajectory summarization and self-critique recover which principles helped or hurt by contrasting successful and failed rollouts of the same problem (difficulty 1).
    \item Local edit proposal keeps each problem's edits small (difficulty 2).
    \item Local-to-global aggregation reconciles all proposals into one compact revision (difficulty 3), much like combining per-example updates into a single batch update with a bounded step size.
\end{itemize}
Crossover complements mutation, which only learns from the mistakes of its own lineage, by transferring principles from prompts that do better on specific problems. Both operators edit the best-performing prompt with a git-style diff, reuse rollouts that RL has already generated, and reflect with the fixed reference model $\pi_{\text{ref}}$ (\autoref{fig:evo-pol-ablation}).

\paragraph{Mutation.}
In each RL iteration of \cref{alg:e-spl}, we mutate the best-performing system prompt,
\begin{align*}
k \leftarrow \arg\max_i V_i, \qquad V_i = \frac{1}{N}\sum_{j=1}^N R(x,y_{i,j}),
\end{align*}
where the returns of all sampled prompts are computed under matched conditions (the same problems and weights). Let $\mathcal{E}_k = \{(x,y_{k,j},r_{k,j})\}_{j=1}^N$ denote the rollouts produced under $\mathbf{s}_k$. The self-reflection prompt $\Psi_{\text{reflect}}$ instructs the reference model to summarize these trajectories and attribute successes and failures to specific reasoning behaviors, producing lessons
\begin{align*}
\ell \sim \pi_{\text{ref}}(\cdot \mid \Psi_{\text{reflect}}, \mathcal{E}_k).
\end{align*}
Given $\ell$, the reference model proposes local edits to the parent prompt for each problem in the batch, $\mathrm{diff} \sim \pi_{\text{ref}}(\cdot \mid \mathbf{s}_k, \ell)$, where each $\mathrm{diff}$ is a list of edit operations. These per-problem edits are aggregated across the batch into a global revision, which is applied to the parent prompt:
\begin{align*}
\mathbf{s}_{\text{mutate}} \leftarrow \mathrm{git.apply}(\mathbf{s}_k, \mathrm{diff}).
\end{align*}
The diff format keeps edits localized, which is especially important as system prompts grow longer. The child is added to the population with its parent's mean rating and increased uncertainty (\mbox{\cref{alg:e-spl}}).

\subsection{Trajectory Summarization in Mutation Self-Reflection}
\label{sec:appendix-trajectory-summary}

The first step of self-reflection prepares the evidence for self-critique, which compares several rollouts of the same problem. Raw rollouts are long reasoning traces in which the decisions that mattered are hard to find, so E-SPL first converts each of them into a structured trajectory summary. For the selected parent prompt $\mathbf{s}_k$, all sampled solutions are grouped by problem instance $x$. Let $\mathcal{R}_x=\{(x,y_{k,j},r_{k,j})\}_{j=1}^{N}$ denote the group of rollouts produced for problem $x$ under $\mathbf{s}_k$. E-SPL only performs self-reflection on problems where the model has both made some mistakes and achieved some success ($0 < V_k < 1$ on problem $x$), so that self-reflection can \emph{contrast} successes and failures under identical instructions; problems that are always solved offer nothing to fix, and problems that are never solved offer no successful behavior to compare against.

For each rollout in a problem-specific group, the reference model $\pi_{\text{ref}}$ is prompted to generate a step-by-step summary describing (i) the reasoning steps taken, (ii) which principles or strategies in the system prompt were (implicitly) used, and (iii) where detours or errors occurred, given the rollout outcome and the ground-truth answer. These summaries preserve the reasoning and its outcome while explicitly highlighting failure modes, producing a compact representation suitable for higher-level self-critique. Recording which principles were used, item (ii), is what allows self-critique to attribute outcomes to specific principles in the next step.

\subsection{Self-Critique in Mutation Self-Reflection}
\label{sec:appendix-local-critique}

Self-critique performs credit assignment at the level of principles. Since all rollouts in a group share the same problem and the same system prompt, the differences between successful and failed attempts point to the reasoning choices that mattered, and from there to principles that were missing, too vague, or misapplied. For each problem $x$, E-SPL constructs a self-critique prompt that jointly presents:
\vspace{-5pt}
\begin{itemize}
    \item multiple trajectory summaries from $\mathcal{R}_x$, labeled as successful or incorrect,
    \item the current system prompt $\mathbf{s}_k$,
    \item the ground-truth answer.
\end{itemize}
The reference model $\pi_{\text{ref}}$ is instructed to analyze these contrasted trajectories to:
\vspace{-5pt}
\begin{enumerate}
    \item identify reasoning patterns that consistently led to success,
    \item diagnose systematic failure modes, and consider additional guidance that would be needed to avoid those failure modes,
    \item determine whether existing principles in the system prompt are insufficient or require further editing.
\end{enumerate}
Based on this analysis, the model proposes structured edit operations (at most $K_{\text{ops}}$ per problem; \S\ref{sec:appendix-max-ops}):
\begin{itemize}
    \item \textbf{add}: introduce a new general principle (or high-level strategy / repeatable workflow) to the system prompt;
    \item \textbf{modify}: refine an existing principle to better encode successful strategies or avoid observed errors.
\end{itemize}
Each proposed principle is designed to express an abstract decision heuristic rather than problem-specific calculations, ensuring transferability across tasks. The reference model is also asked to state when each principle applies and which problem features trigger it, which is why many learned principles take the form ``for problems of type $X$, do $Y$'' (e.g., \autoref{fig:examples-principles}).

\subsection{Local Edit Proposal}
\label{sec:appendix-max-ops}

The patch synthesis stage turns these diagnoses into edits of the prompt, starting with a limit on how much any single problem can change it. The edit operations proposed by self-critique for a problem constitute its local edit proposal. To stabilize evolution and prevent overly aggressive prompt drift, E-SPL restricts each per-problem proposal to at most $K_{\text{ops}}$ edit operations, where $K_{\text{ops}}$ is a tunable hyperparameter ($K_{\text{ops}}=2$ in our experiments). This encourages incremental knowledge accumulation and localized corrections driven by concrete failures; empirically, bounding the number of local edits leads to steady refinement of the principles in the system prompt while preserving useful structure learned in earlier generations. Small edits also suit RL training: since the policy is trained under the evolving prompts, they keep the prompts it sees from changing abruptly, and they make a child's rating attributable to a few specific changes.

\subsection{Local-to-Global Aggregation}
\label{sec:appendix-global-revision}

In each RL iteration, multiple problems may independently propose local edits. Directly applying all of them can lead to conflicting updates, redundancy, or uncontrolled growth of the principle set in the system prompt. E-SPL therefore aggregates the local edit proposals into a global revision plan. Proposed additions are first included as candidate principles; then, given the selected system prompt $\mathbf{s}_{k}$, these candidates, and the proposed modifications, the reference model is prompted to produce two types of operations:
\begin{itemize}
    \item \textbf{modify}: refine an existing principle by aggregating all edits proposed to it across the batch, incorporating repeated evidence;
    \item \textbf{merge}: replace several semantically overlapping principles with a unified higher-level strategy or a more universal rule across domains.
\end{itemize}
All final principles in the system prompt are encouraged to be short (at most 32 words), abstract, and strategy-oriented, which prevents uncontrolled growth of the system prompt and yields a compact declarative knowledge base that evolves steadily throughout training. The revised principles form the mutated child $\mathbf{s}_{\text{mutate}}$, which enters the population with its parent's mean rating and increased uncertainty (\S\ref{sec:appendix-trueskill}).

\subsection{Crossover}
\label{sec:appendix-crossover-operator}

\begin{algorithm}[t]
\caption{$\mathrm{Crossover}(\mathcal{B}, R, \Omega, \pi_{\theta}, \pi_{\text{ref}} )$}
    \label{alg:crossover}
    \begin{algorithmic}
        \STATE {\bfseries Input:} List of $B$ problems: $\mathcal{B}$; reward function $R(\cdot, \cdot)$.
        \STATE {\bfseries Input:} List of $M$ system prompts: $\Omega$.
        \STATE {\bfseries Input:} RL policy $\pi_{\theta}(\cdot)$, reference policy $\pi_{\text{ref}}(\cdot)$.
        \vspace{5pt}

        \STATE Let $\Phi \in \mathbb{R}^{M \times B}$ \hfill \COMMENT{{\color{Brown}$\Phi_{i,b}$ is the empirical return of the $i$-th system prompt in $\Omega$ on the $b$-th problem in $\mathcal{B}$}}
        \FOR{$(\mathbf{s}_{i}, \mu_i, \sigma_i)$ $\textbf{in}$ $\Omega$}
            \FOR{$x_{b}$ $\textbf{in}$ $\mathcal{B}$}
                \STATE Sample $\{y_{i, b, j}\}_{j=1}^{N}$ from $\pi_{\theta}$: $y_{i, b, j}\sim \pi_{\theta}(\cdot \mid \mathbf{s}_{i}, x_{b})$ \hfill \COMMENT{{\color{Brown}In practice, {reuse} RL rollouts (Alg \ref{alg:e-spl})}}
                \STATE $\Phi_{i, b} = \frac{1}{N} {\sum}_{j=1}^{N} R(x_{b}, y_{i, b, j})$
            \ENDFOR
        \ENDFOR
        \STATE $k\leftarrow \mathrm{argmax}_{i} {\sum}_{b} \Phi_{i,b}$ \hfill \COMMENT{{\color{Brown}$\mathbf{s}_{k}$ is the overall top prompt}}%
        \vspace{5pt}

        \STATE $\boldsymbol{\varphi} = [\,]$ \hfill \COMMENT{{\color{Brown}Problems each $\mathbf{s}_{i}$ was the top prompt for}}
        \FOR{$(\mathbf{s}_{i}, \mu_i, \sigma_i)$ $\textbf{in}$ $\Omega$}
            \STATE $\rho_{i}=[\,]$
            \FOR{$x_{b}$ $\textbf{in}$ $\mathcal{B}$}
                \IF{$i=\mathrm{argmax}_{z} \Phi_{z,b}$}
                    \STATE $\rho_{i}$.$\mathrm{append}(x_{b})$ \hfill \COMMENT{{\color{Brown} $\mathbf{s}_{i}$ is the top prompt for $x_{b}$}}
                \ENDIF
            \ENDFOR
            \STATE $\boldsymbol{\varphi}.\mathrm{append}((\mathbf{s}_{i}, \rho_{i}))$ \hfill \COMMENT{{\color{Brown}Analyze why $\mathbf{s}_{i}$ did best on $\rho_{i}$}}
        \ENDFOR
        \STATE Create {\color{magenta}\textbf{{\text{crossover}} self-reflection}} prompt $\Psi_{\text{cr}}$ using $\boldsymbol{\varphi}$.
        \vspace{5pt}

        \STATE $\mathrm{diff}_{\text{cr}} \sim \pi_{\text{ref}}(\cdot \mid \mathbf{s}_{k}, \Psi_{\text{cr}})$
        \STATE $\mathbf{s}_{\text{crossover}} \leftarrow \mathrm{git}.\mathrm{apply}(\mathbf{s}_{k}, \mathrm{diff}_{\text{cr}})$ \hfill  \COMMENT{{\textbf{Crossover step}}}%
        \vspace{5pt}

        \STATE $\mu' = ({\sum}_{i}\frac{\mu_{i}}{\sigma_{i}^{2}})/({\sum}_{i}\frac{1}{\sigma_{i}^{2}})$, $\sigma' = \sqrt{{{1}/({{\sum}_{i}{{1}/{\sigma_{i}^{2}}}})}}$

        \STATE \textbf{Return} $(\mathbf{s}_{\text{crossover}}, \mu', \sigma')$
    \end{algorithmic}
\end{algorithm}

Crossover complements mutation by transferring strengths across lineages into the best prompt (\S\ref{sec:appendix-self-reflection}). The crossover operator (\cref{alg:crossover}) differs from mutation in the evidence used for self-reflection: instead of contrasting the rollouts of a single prompt, it contrasts the problem-wise performance of multiple prompts. For a batch of problems $\mathcal{B}$, we compute the empirical return matrix
\begin{align*}
\Phi_{i,b} = \frac{1}{N}\sum_{j=1}^N R(x_b,y_{i,b,j})
\end{align*}
by reusing rollouts already generated by RL. For each system prompt $\mathbf{s}_{i}$, we identify the problems $\rho_{i}$ in $\mathcal{B}$ on which $\mathbf{s}_{i}$ performed best, yielding a list $\boldsymbol{\varphi}$ of prompt-specific wins, from which we construct the crossover reflection prompt $\Psi_{\text{cr}}$ that contrasts the differential strengths of the parent prompts. The reference model then proposes edits to the overall top-performing prompt $\mathbf{s}_k$:
\begin{align*}
\mathrm{diff}_{\text{cr}} \sim \pi_{\text{ref}}(\cdot \mid \mathbf{s}_k, \Psi_{\text{cr}}), \qquad
\mathbf{s}_{\text{crossover}} \leftarrow \mathrm{git.apply}(\mathbf{s}_k, \mathrm{diff}_{\text{cr}}),
\end{align*}
producing a child prompt that selectively recombines complementary instruction segments from multiple parents. The child is initialized with the precision-weighted combination of its parents' ratings. Because a single win can be due to chance, the reference model is asked to judge whether each win is explained by specific principles, and to preserve the principles of $\mathbf{s}_k$ that explain its own wins. Crossover is applied with probability $p_{\text{crossover}}$ in each iteration (\S\ref{sec:appendix-hyperparameters}), in addition to mutation.

\subsection{Hyperparameters}
\label{sec:appendix-hyperparameters}

The hyperparameters below control the components described in this section and the fitness model in \S\ref{sec:appendix-trueskill}. We use the following hyperparameters:
\begin{itemize}
    \item LoRA \citep{lora} rank of 32, learning rate $\alpha=4\times10^{-5}$, using the Adam optimizer \citep{kingma2014adam} with $(\beta_1, \beta_2)=(0.9, 0.95)$.
    \item Batch size of 10 problems, $M=3$ system prompts per batch, and group size $N=5$ rollouts per problem and system prompt.
    \item Sliding window of size $K=10$: selection is restricted to the $K$ most recently created prompts in the population $\mathcal{S}$, to avoid selecting stale prompts.
    \item Uncertainty bonus for children of the mutation operator: $\Delta\sigma=1$; exploration weight in the selection score: $\lambda=2.0$; crossover probability: $p_{\text{crossover}}=0.2$.
    \item TrueSkill (all defaults): $\mu_{0}=25$, $\sigma_0=25/3$, performance noise $\beta=25/6$, drift $\tau=25/300$, draw probability $p_{\text{draw}}=0.1$.
    \item We find that a simplified selection strategy is quite effective: always add the system prompt with the highest score ($\mathrm{score}_{i} = \mu_{i}+\lambda \sigma_{i}$) to the batch, and randomly sample the remaining prompts from the sliding window.
    \item $K_{\text{ops}}=2$, the maximum number of local edits per problem in the mutation operator.
\end{itemize}

\subsection{Root System Prompt for Agentic Search}
\label{sec:appendix-root-prompt}

For agentic search, evolution starts from the following root system prompt:
\begin{runcard}[title={Root System Prompt for Agentic Search}, breakable, halign=left]
You are an expert assistant who solves tasks using a Wikipedia search tool.

\smallskip
Here are instructions for how to solve a problem:
\begin{enumerate}
    \item Think step by step before calling the tool and after you receive the result of the tool call. Decide what queries to call the tool with.
    \item Call the tool with the queries you have decided on.
    \item Think step by step again after you receive the result of the tool call. If you have the information you need, you can stop here.
    \item Otherwise, come up with new queries that combine information from the previous results.
    \item Include your final answer after the ``Answer:'' prefix. The answer should be between one to five words.
\end{enumerate}

\smallskip
Here is an example of solving a real question:

``Between 2020 and 2025, which year did New York City see the most population growth and how did San Francisco population change in that year?''
\begin{enumerate}
    \item Think step by step: In order to answer this question, I need to know the population of New York City and San Francisco between 2020 and 2025. I will search for the population of New York City in each year
    \item Calling search tool: \texttt{<tool\_call>\{"name": "search", "arguments": \{"query\_list": ["Population New York city between 2020 and 2025"]\}\}</tool\_call>} (Output omitted for brevity)
    \item Think step by step again: I have the population of New York City in each year, and I see that the population of New York City grew the most in 2024. I need to know the population of San Francisco in 2024. I will search for the population of San Francisco in each year.\\
    \texttt{<tool\_call>\{"name": "search", "arguments": \{"query\_list": ["Population San Francisco between 2023 and 2024"]\}\}</tool\_call>} (Output omitted for brevity)
    \item Answer: The population of New York City grew the most in 2024, and the population of San Francisco changed by XXXX in 2024.
\end{enumerate}
\end{runcard}
\medskip

E-SPL evolves this root prompt into the system prompt given in full in \S\ref{sec:full-prompt-agentic-search}; \autoref{fig:prompt-evolution-search} compares the two (abridged).

\section{TrueSkill-Based Fitness Modeling for E-SPL}
\label{sec:appendix-trueskill}

To aggregate noisy batch-wise prompt comparisons into persistent competence estimates, we adopt the Bayesian skill rating model \textbf{TrueSkill} \citep{trueskill}. Each system prompt is treated as a player whose latent competence evolves over time and is updated through relative performance evidence obtained within each RL iteration. We use the default TrueSkill hyperparameters $\mu_0=25$, $\sigma_0=25/3$, performance noise $\beta=25/6$, drift $\tau=25/300$, and draw probability $p_{\text{draw}}=0.1$. For completeness, we review the mathematical details of this {off-the-shelf} TrueSkill system below.

\paragraph{Overview.}
Within an RL iteration, all $M$ sampled prompts are evaluated on the same problems with the same weights, so their returns $V_i$ are directly comparable; across iterations, both the problems and the weights change, so raw returns are not. E-SPL therefore treats each iteration as a small tournament among the $M$ prompts and only uses their ranking. TrueSkill aggregates these noisy rankings into a persistent Gaussian belief $\mathcal{N}(\mu_i,\sigma_i^2)$ over the latent skill of each prompt: $\mu_i$ estimates how good prompt $\mathbf{s}_i$ is, and $\sigma_i$ quantifies the uncertainty of this estimate. Winning against highly rated opponents raises $\mu_i$ more than winning against lowly rated ones, and $\sigma_i$ shrinks as a prompt takes part in more comparisons. Selection uses the optimistic score $\mathrm{score}_i=\mu_i+\lambda\sigma_i$ with $\lambda=2.0$, an upper confidence bound on skill \citep{srinivas2009gaussian, kaufmann2012bayesian, lucb, zhu2024ucb}: prompts with high estimated skill are exploited, while new or rarely compared prompts, whose $\sigma_i$ is still large, are explored.

\paragraph{Skill and performance model.}
Each system prompt $\mathbf{s}_i$ is associated with a Gaussian skill variable $z_i \sim \mathcal{N}(\mu_i,\sigma_i^2)$, where $\mu_i$ denotes estimated competence and $\sigma_i$ its uncertainty. The latent performance of $\mathbf{s}_i$ in an iteration is modeled as a noisy skill, $p_i \sim \mathcal{N}(z_i,\beta^2)$, so that the marginal performance variance is $\mathrm{Var}(p_i)=\sigma_i^2+\beta^2$. Before each rating update, we apply Gaussian drift $\tau$ in skill space to model temporal uncertainty growth: $\sigma_i^2 \leftarrow \sigma_i^2 + \tau^2$.

\paragraph{Ranking observations and draw margin.}
Within each RL iteration, matched rollouts yield scalar returns $\{V_i\}_{i=1}^{M}$ for selected prompts. Sorting these values produces a ranking from best to worst. Ties are encoded by an integer array $\texttt{ties}$ such that adjacent positions $i$ and $i+1$ are treated as a draw if $\texttt{ties}[i]=\texttt{ties}[i+1]$, and otherwise as a win for position $i$ over $i+1$. This yields a chain of $M-1$ adjacent pairwise constraints. We compute a global draw margin in performance-difference units, $\epsilon_{\mathrm{abs}} = \Phi^{-1}\big((p_{\text{draw}}+1)/2\big)\sqrt{2}\,\beta$, where $\phi(\cdot)$ and $\Phi(\cdot)$ denote the pdf and cdf of the standard normal distribution. For each pairwise comparison with total standard deviation $c$, inference uses the normalized margin $\epsilon = \epsilon_{\mathrm{abs}}/c$ and the normalized performance difference $t=(\mu_i-\mu_j)/c$. With $d_{\pm}=\pm\epsilon-t$, the one-sided (win/loss) and two-sided (draw) correction terms are
\begin{align*}
v_{\text{win}} &= \frac{\phi(t-\epsilon)}{\Phi(t-\epsilon)}, &
w_{\text{win}} &= v_{\text{win}}\,(v_{\text{win}}+t-\epsilon), \\
v_{\text{draw}} &= \frac{\phi(d_-)-\phi(d_+)}{\Phi(d_+)-\Phi(d_-)}, &
w_{\text{draw}} &= v_{\text{draw}}^2+\frac{d_+\,\phi(d_+)-d_-\,\phi(d_-)}{\Phi(d_+)-\Phi(d_-)}.
\end{align*}

\paragraph{Approximate inference for full rankings.}
For rankings with more than two prompts, we perform EP-style iterative message passing \citep{minka2001family} over the adjacent constraints in performance space. Constraint $k$ connects the prompts at positions $i=k$ and $j=k+1$, and sends each of them a Gaussian message in natural parameters (precision $\pi$ and precision-adjusted mean $\eta$), such as $(\pi_{k\to i},\eta_{k\to i})$, initialized to zero. The performance-space prior of prompt $\mathbf{s}_i$ is $\mathcal{N}(\mu_i,\sigma_i^2+\beta^2)$, with natural parameters $\pi_i^{(0)}=1/(\sigma_i^2+\beta^2)$ and $\eta_i^{(0)}=\mu_i\,\pi_i^{(0)}$. Each update of constraint $k$ has three steps:
\begin{enumerate}
    \item \emph{Cavity.} Combine the prior with all messages except the one from constraint $k$:
    \begin{align*}
    \bar\pi_i=\pi_i^{(0)}+\sum_{k'\neq k}\pi_{k'\to i}, \qquad
    \bar\eta_i=\eta_i^{(0)}+\sum_{k'\neq k}\eta_{k'\to i},
    \end{align*}
    giving $\bar\mu_i=\bar\eta_i/\bar\pi_i$ and $\bar\sigma_i^2=1/\bar\pi_i$; likewise for $j$.
    \item \emph{Moment update.} With $c^2=\bar\sigma_i^2+\bar\sigma_j^2$, $t=(\bar\mu_i-\bar\mu_j)/c$, and $\epsilon=\epsilon_{\mathrm{abs}}/c$, let $(v,w)$ be the draw correction terms if the pair is a draw and the win terms otherwise. Then
    \begin{align*}
    \hat\mu_i=\bar\mu_i+\frac{\bar\sigma_i^2}{c}\,v, \qquad
    \hat\sigma_i^2=\bar\sigma_i^2\Big(1-\frac{\bar\sigma_i^2}{c^2}\,w\Big),
    \end{align*}
    and analogously for $j$, with $-v$ in place of $v$.
    \item \emph{Message update.} The new message is the difference between the updated and cavity natural parameters, with the precision clamped to be nonnegative for numerical stability: $\pi_{k\to i}=\max(0,\,1/\hat\sigma_i^2-\bar\pi_i)$ and $\eta_{k\to i}=\hat\mu_i/\hat\sigma_i^2-\bar\eta_i$; likewise for $j$.
\end{enumerate}
These updates are iterated over all constraints for a fixed number of sweeps or until the messages change by less than a small threshold.

\paragraph{Mapping back to skill space.}
After convergence, the total message that prompt $\mathbf{s}_i$ receives from all constraints is $\pi_{\text{msg}}=\sum_k \pi_{k\to i}$ and $\eta_{\text{msg}}=\sum_k \eta_{k\to i}$. When $\pi_{\text{msg}}>0$, this message is mapped through the performance noise $\beta$ and combined with the skill prior $\mathcal{N}(\mu_i,\sigma_i^2)$ after drift:
\begin{align*}
\pi_i^{\text{skill}}=\frac{1}{\sigma_i^2}+\frac{\pi_{\text{msg}}}{1+\beta^2\pi_{\text{msg}}}, \qquad
\eta_i^{\text{skill}}=\frac{\mu_i}{\sigma_i^2}+\frac{\eta_{\text{msg}}}{1+\beta^2\pi_{\text{msg}}},
\end{align*}
giving the updated rating $\sigma_i^{2}=1/\pi_i^{\text{skill}}$ and $\mu_i=\eta_i^{\text{skill}}/\pi_i^{\text{skill}}$.

\paragraph{Initialization for mutation and crossover.}
A mutated child prompt inherits its parent's mean while inflating uncertainty: $\mu_{\text{child}}=\mu_k$ and $\sigma_{\text{child}}=\sqrt{\sigma_k^2+\Delta\sigma^2}$. For crossover among parents $\{i\}$, we perform precision-weighted fusion:
\begin{align*}
\mu'=\frac{\sum_i \mu_i/\sigma_i^2}{\sum_i 1/\sigma_i^2}, \qquad
\sigma'^2=\frac{1}{\sum_i 1/\sigma_i^2} .
\end{align*}
\paragraph{Summary.} Together, these components turn noisy, iteration-specific comparisons into a persistent estimate of each prompt's fitness. Each RL iteration ranks only the $M$ prompts it sampled, on one batch and under the current weights; TrueSkill folds this local comparative evidence into the ratings of those prompts, the drift $\tau$ lets ratings adapt as the policy changes, and new children enter with ratings inherited from their parents. The ratings thus accumulate into a coherent evolutionary fitness landscape over the whole population, which selection explores through the optimistic score $\mu_i+\lambda\sigma_i$.

\clearpage
\newpage

\section{Additional Results and Discussion}
\label{sec:appendix-additional-results}

This section complements the main results with a compute-matched RL baseline and results across random seeds (\S\ref{sec:appendix-compute-matched}), sample efficiency (\S\ref{sec:appendix-sample-efficiency}), a comparison with sequential evolution-then-RL training (\S\ref{sec:appendix-sequential}), the effect of crossover (\S\ref{sec:appendix-crossover}), and computational cost (\S\ref{sec:appendix-cost}), followed by qualitative examples (\S\ref{sec:appendix-search-prompt}--\ref{sec:appendix-examples}), a discussion of limitations (\S\ref{sec:appendix-discussion}), and open questions and future directions (\S\ref{sec:appendix-future}).

\subsection{Compute-Matched RL Baseline and Variance Across Seeds}
\label{sec:appendix-compute-matched}

In the main experiments, E-SPL samples $N=5$ rollouts under each of $M=3$ system prompts per problem ($M \times N = 15$ rollouts), whereas the RL-only baselines sample $N=5$ rollouts under a single fixed system prompt. To control for this difference, we train a \textbf{compute-matched} GRPO baseline with group size $N=15$, which matches the RL component of E-SPL in the number of sampled tokens, policy rollouts, and gradient updates, and in wall-clock time; the cost of self-reflection is analyzed separately in \S\ref{sec:appendix-cost}. Apart from the group size of this baseline, all methods share the same RL hyperparameters (batch size, learning rate, optimizer, and LoRA rank; \S\ref{sec:appendix-hyperparameters}), so the gains of E-SPL are not due to differences in RL-specific hyperparameters.

As shown in \autoref{tab:compute-matched}, increasing the GRPO group size without increasing system prompt diversity yields only marginal gains ($+0.5$ on HMMT and $+0.4$ on BeyondAIME), while E-SPL improves over the compute-matched baseline by $2.2$ and $5.9$ points, respectively, with the largest gain in easy-to-hard generalization. To assess run-to-run variability, we also repeat E-SPL on AIME $\rightarrow$ BeyondAIME with 3 random seeds, obtaining $44.8 \pm 0.4$ (mean $\pm$ standard deviation), well above both the compute-matched GRPO baseline ($39.2$) and evolution alone ($40.0$).

\begin{table}[H]
\centering
\small
\caption{Compute-matched RL baseline (success rate, \%). ``RL (GRPO), compute-matched'' uses group size $15$, matching E-SPL's $M \times N = 3 \times 5$ rollouts per problem.}
\label{tab:compute-matched}
\vspace{5pt}
\begin{tabular}{lcc}
\toprule
Method & HMMT 23/24 $\rightarrow$ 25 & AIME $\rightarrow$ BeyondAIME \\
\midrule
Base model & 39.3 & 36.5 \\
Self-Reflection & 47.0 & 39.1 \\
Evolution & 48.6 & 40.0 \\
RL (CISPO) & 47.3 & 38.4 \\
RL (GRPO) & 50.0 & 38.8 \\
RL (GRPO), compute-matched & 50.5 & 39.2 \\
\textbf{Evolution + RL (E-SPL)} & \textbf{52.7} & \textbf{45.1} \\
\bottomrule
\end{tabular}
\end{table}

\subsection{Sample Efficiency}
\label{sec:appendix-sample-efficiency}

We measure sample efficiency as a baseline's total number of training epochs divided by the number of epochs E-SPL needs to reach that baseline's final performance (\autoref{tab:sample-efficiency}). E-SPL reaches the final performance of GRPO with $2.1\times$ fewer epochs on HMMT and $5.9\times$ fewer on BeyondAIME, and that of the compute-matched GRPO baseline with $1.8\times$ and $2.6\times$ fewer epochs, respectively.

\begin{table}[H]
\centering
\small
\caption{Sample-efficiency gain of E-SPL: a baseline's total number of training epochs divided by the number of epochs E-SPL needs to reach that baseline's final performance.}
\label{tab:sample-efficiency}
\vspace{5pt}
\begin{tabular}{lcc}
\toprule
E-SPL vs. & HMMT 23/24 $\rightarrow$ 25 & AIME $\rightarrow$ BeyondAIME \\
\midrule
RL (GRPO) & $2.1\times$ & $5.9\times$ \\
RL (GRPO), compute-matched & $1.8\times$ & $2.6\times$ \\
\bottomrule
\end{tabular}
\end{table}

\subsection{Joint versus Sequential Evolution and RL}
\label{sec:appendix-sequential}

A natural alternative to joint training is to evolve system prompts first and run RL afterwards. We run one epoch of prompt evolution (the same evolutionary algorithm as E-SPL, without gradient updates) on DAPO100, followed by RL without further prompt evolution. As shown in \autoref{tab:sequential}, this sequential variant improves over evolution and RL alone, but remains below E-SPL: interleaving evolution and RL has clear benefits, because evolution always reflects on the latest failure modes of the current policy. Note that sequential training is difficult to compare against E-SPL in a strictly compute-matched regime, since prompt evolution itself requires many policy rollouts.

\begin{table}[H]
\centering
\small
\caption{Joint versus sequential training (success rate, \%). ``Evolution, then RL'' runs one epoch of prompt evolution without gradient updates, followed by RL.}
\label{tab:sequential}
\vspace{5pt}
\begin{tabular}{lc}
\toprule
Method & DAPO100 $\rightarrow$ AIME 25 \\
\midrule
Base model & 48.6 \\
Evolution & 57.3 \\
RL (GRPO) & 56.3 \\
Evolution, then RL & 58.1 \\
\textbf{Evolution + RL (E-SPL)} & \textbf{60.6} \\
\bottomrule
\end{tabular}
\end{table}

\subsection{Effect of Crossover}
\label{sec:appendix-crossover}

\autoref{tab:crossover} compares E-SPL with mutation only and with both mutation and crossover (\S\ref{sec:appendix-crossover-operator}). The two variants reach comparable final performance, and both outperform RL alone.

\begin{table}[H]
\centering
\small
\caption{Effect of crossover (success rate, \%). ``Mutation + RL'' is E-SPL without the crossover operator.}
\label{tab:crossover}
\vspace{5pt}
\begin{tabular}{lcc}
\toprule
Method & HMMT 23/24 $\rightarrow$ 25 & AIME $\rightarrow$ BeyondAIME \\
\midrule
RL (GRPO) & 50.0 & 38.8 \\
Mutation + RL & 51.6 & 45.1 \\
Mutation + Crossover + RL & 52.7 & 44.5 \\
\bottomrule
\end{tabular}
\end{table}

\subsection{Computational Cost}
\label{sec:appendix-cost}

E-SPL evaluates system prompts by reusing the rollouts already generated for RL, so it requires no additional environment interactions or held-out evaluation. Its additional cost is LLM self-reflection by the reference model for mutation and crossover, which only requires a sampling server. In our current implementation, E-SPL adds about $20\%$ wall-clock time per training iteration, mostly because self-reflection is synchronous: the next RL iteration begins only after the self-reflection of the current iteration has finished. In principle, self-reflection can be fully parallelized with RL sampling and gradient updates.

\subsection{Discovered System Prompt for Agentic Search}
\label{sec:appendix-search-prompt}

\autoref{fig:prompt-evolution-search} shows the system prompt for the agentic search task before and after evolution with E-SPL (abridged; the full evolved system prompt is given in \S\ref{sec:full-prompt-agentic-search}).

\subsection{Additional Examples of E-SPL Model Behavior}
\label{sec:appendix-examples}

\autoref{fig:example-e-spl-1} and \autoref{fig:example-e-spl-2} provide detailed examples of how the RL model learns to utilize principles from learned system prompts to solve challenging math problems. \autoref{fig:example-e-spl-3} and \autoref{fig:example-e-spl-4} show further successful examples, while \autoref{fig:example-e-spl-6} and \autoref{fig:example-e-spl-5} illustrate failure modes.

\subsection{Discussion and Limitations}
\label{sec:appendix-discussion}

\paragraph{Fixed reference model for self-reflection.} E-SPL performs self-reflection with a fixed reference model $\pi_{\text{ref}}$, which works better than using the continually changing RL policy (\autoref{fig:evo-pol-ablation}). This is analogous to using a fixed LLM judge to provide rewards during RL: although its weights are fixed, the reference model always reflects on the latest on-policy trajectories and aims to help the online policy achieve higher returns. The RL policy is trained to execute the task as well as possible, not to improve self-reflection; better weights lead to better system prompts by giving the reference model better trajectories to reflect upon. A limitation is that the reference model must be sufficiently capable of self-reflection and instruction following for E-SPL to work; this holds for the latest generation of open-weight models, but may not hold for weaker ones.

\paragraph{Declarative and procedural knowledge.} We use these terms to describe our intuition and the qualitative patterns we observe, rather than a directly measured mechanism. \emph{Declarative knowledge} refers to strategies and principles that can be verbalized, and can therefore be placed in a system prompt; \emph{procedural knowledge} refers to behavior changes that cannot be fully verbalized. Concretely, after an RL update $\theta \rightarrow \theta'$ under system prompt $\mathbf{s}_{\text{old}}$, however strong the evolution process is for a new system prompt $\mathbf{s}_{\text{new}}$, a non-zero gap $D_{\mathrm{KL}}\big(p(\cdot \mid \theta, \mathbf{s}_{\text{new}}) \,\|\, p(\cdot \mid \theta', \mathbf{s}_{\text{old}})\big)$ remains; this gap measures the procedural knowledge stored in the weight update.

\paragraph{System prompt diversity.} The population starts from a single root system prompt, so all diversity in system prompts is generated by E-SPL itself. In our experiments, the bottleneck is usually not the population collapsing to semantically similar system prompts, as the base models are capable enough to extract novel lessons from new mistakes; rather, it is self-reflection occasionally arriving at wrong conclusions or strategies, which then requires further rounds of self-correction during evolution or human oversight of the evolved system prompts.

\subsection{Open Questions and Future Directions}
\label{sec:appendix-future}

E-SPL lets a model learn from its own experience in two places: RL updates its weights, while self-reflection updates an experiential context in the form of system prompts. This builds on the in-context learning and instruction-following abilities of base LLMs, which RL alone does not fully exploit: RL learns by trial and error under a fixed context, without letting the model revise its own instructions. We see several open questions in using evolutionary self-rewrite to improve RL training.

\paragraph{What belongs in the context, and what in the weights?} RL outperforms prompt evolution on agentic search but is on par with it on math reasoning (\S\ref{sec:main-results}). It is hard to predict in advance which will help more on a new domain, since the learning bottleneck can be any mix of declarative and procedural knowledge. A better understanding of this split could guide how computation is allocated between context updates and weight updates.

\paragraph{Consolidating context into weights.} The RL-trained model currently depends on the evolved system prompts for its best performance, yet some principles may become routine for the policy as training proceeds, much as declarative knowledge becomes procedural with practice in human skill acquisition \citep{anderson1982acquisition}. Such principles could be identified with E-SPL's own ratings, by checking whether deleting one no longer lowers a child's rating, and then moved into the weights to keep the system prompt compact, for example through context distillation \citep{askell2021general, snell2022learning}. Which knowledge should remain explicit, where it can be inspected and corrected (\autoref{fig:corrected-principles}), and which can move into the weights is an open question.

\paragraph{Self-rewrite as an RL problem.} In E-SPL, self-rewrite (reflection, mutation, and crossover) is carried out by a fixed reference model, which is more stable than using the changing policy (\autoref{fig:evo-pol-ablation}) but bounds the quality of the edits. A natural extension is to treat self-rewrite itself as an RL problem: the model would be trained not only to solve tasks under its system prompts, but also to rewrite those system prompts, with the rating gain of each new system prompt over its parent as a reward. Doing so stably, without the drift we observe when the editor changes alongside the policy, remains an open question.

\begin{figure}[p]
\centering

\caption{\centering
\textbf{An example of the system prompt discovered by E-SPL for RL with search engines.} Here, the discovered system prompt specifies a workflow that includes: {\color{Brown}generative self-verification, detailed instructions on tool use, critical formatting rules, heuristics for determining the reliability of various information sources, actionable guidance for avoiding common failure modes}, etc.
}
\label{fig:prompt-evolution-search}

\begin{runcard}[title={System Prompt for RL Agent with Search Engine (OpenAI gpt-oss-120b)}]

\begin{runbox}[title={System Prompt \emph{Before} E-SPL}, halign=left]

You are an expert assistant who solves tasks using a Wikipedia search tool. \quad {\color{gray}$\cdots$}

{Here is an example of solving a real question:}

{
``Between 2020 and 2025, which year did New York City see the most population growth and how did San Francisco population change in that year?''
} \quad {\color{gray}$\cdots$}

{
Calling search tool:
\texttt{<tool\_call>\{"name": "search", "arguments": \{"query\_list": ["Population New York City between 2020 and 2025"]\}\}</tool\_call>}
}

{\quad $\cdots$}

\end{runbox}

\begin{runbox}[title={\textbf{System Prompt \emph{After} E-SPL}}]
\frenchspacing

You are an expert research assistant. \quad $\cdots$
\quad
{\color{Brown}Workflow (follow \emph{exactly} in this order) \quad $\boldsymbol{\cdots}$}

\smallskip
\textbf{3. Call the search tool once} --- Use \textbf{exactly} this syntax \quad $\cdots$

\smallskip
\textbf{4. Verification block} --- For each result returned,
{\textcolor{Brown}{output the following four lines in this order and nothing else}}:

\smallskip
\texttt{Source: <domain>} \\
\texttt{Summary: <one-sentence summary>} \\
\texttt{Status: VERIFIED (or NOT VERIFIED)} \\
\texttt{Phrase: "<exact phrase>"} \ \ (only if Status is VERIFIED)

\smallskip
{\textcolor{Brown}{A result is VERIFIED only if:}}

\begin{itemize}
  \item the exact target phrase appears verbatim in the snippet, \textbf{and}
  \item the domain ends with \texttt{.org}, \texttt{.edu}, \texttt{.gov}, is Wikipedia,
  \textbf{or} a major reputable outlet (e.g., \texttt{nytimes.com}, \texttt{bbc.com}, \texttt{reuters.com}, \texttt{theguardian.com})
\end{itemize}

If the domain is not on this list, mark the result \textbf{NOT VERIFIED} even if the phrase is present.

\smallskip
\textbf{5. Refine query (optional)} ---
{\textcolor{Brown}{If all results are NOT VERIFIED, modify the query and repeat Step 3.}}

\smallskip
\textbf{Critical formatting rules (must be obeyed)}

\begin{itemize}
  \item All tool calls must be wrapped exactly as \texttt{<tool\_call>\{...\}</tool\_call>}
  \item The only user-visible output after verification is the single \texttt{Answer:} line; {{no bullet points, markdown, blank lines, or extra text}}
\end{itemize}

\textbf{Positive example (correct flow)}

\smallskip
\texttt{User: "What other name is used for the 4 x 400 m relay?"} \\
\texttt{Query: "4 x 400 metres relay"} \\
\texttt{<tool\_call>\{"name": "search", "arguments": \{"query\_list": ["4 x 400 metres relay"]\}\}</tool\_call>} \\
\texttt{Source: en.wikipedia.org} \\
\texttt{Summary: The 4 x 400 metres relay is also known as the "long relay".} \\
\texttt{Status: VERIFIED} \\
\texttt{Phrase: "long relay"} \quad $\cdots$

\smallskip
{\textcolor{Brown}{
Follow exactly this loop for every user query.
Your objective is to prove the answer with a verbatim phrase from a reliable source
before you \emph{ever} write the final \texttt{Answer:} line.
}}

\end{runbox}
\end{runcard}

\end{figure}

\begin{figure}[p]
\centering
\caption{\textbf{An example of how the RL model learns to utilize learned system prompts} (numbered G1~$\dots$~G30) \textbf{to solve problems.}}

\begin{runcard}[title={Example of How E-SPL Shapes Model Behavior}]

\smallskip

\begin{runbox}[title={\textbf{Learned System Prompt}}]
When solving math problems, you MUST first carefully read and understand the helpful instructions and principles below:

{\color{Brown}\hl{{[G1]}}. When optimizing sums, set parameters to cancel dominant terms. For quadratic forms, choose parameters that minimize the leading coefficients, reducing the problem to analyzing remainder terms.}

$\cdots$

{\color{Brown}\hl{{[G22]}}. In modular arithmetic and sums, decompose terms via remainders or moduli. Use periodicity for remainder sums and CRT for coprime factors, ensuring multiplicative solutions for square-free moduli.}

$\cdots$
\end{runbox}

\begin{runbox}[title={Problem}]
Let \(\mathbb{Z}/n\mathbb{Z}\) be the set of all integers modulo \(n\) (hence \(\mathbb{Z}/n\mathbb{Z}\) contains \(n\) elements). Find the second smallest composite number \(n\), such that there exists a bijective function \(g:\mathbb{Z}/n\mathbb{Z}\to \mathbb{Z}/n\mathbb{Z}\) such that the \(101\) functions
\[ g(x),\ g(x)+x,\ g(x)+2x,\ \cdots,\ g(x)+100x \]
are all bijective from \(\mathbb{Z}/n\mathbb{Z}\) to \(\mathbb{Z}/n\mathbb{Z}\).
\end{runbox}

\begin{runbox}[title={Model Output}]
\textbf{Understanding the Condition:}
Each function \(h_k(x) = g(x) + kx\) (for \(k=0,1,\ldots,100\)) must be a permutation of \(\mathbb{Z}/n\mathbb{Z}\). Since \(g\) is bijective, \(h_0(x)=g(x)\) is bijective by assumption. For \(k \geq 1\), \(h_k\) is bijective if and only if for every \(a \in \mathbb{Z}/n\mathbb{Z}\), the equation \(g(x) + kx = a\) has a unique solution \(x\).

\textbf{Alternatively}, since \(g\) is a permutation, we can set \(y = g(x)\) and then \(x = g^{-1}(y)\), so
$h_k(g^{-1}(y)) = y + k g^{-1}(y).$
{But perhaps} a better approach is to consider the differences.
\textbf{Note:}
\(h_k\) is bijective if and only if the map \(x \mapsto g(x) + kx\) is injective. That is, if
$g(x) + kx = g(y) + ky$
then \(x = y\). Equivalently, $g(x) - g(y) = -k(x-y).$

\textbf{Another idea}: Consider the composition \(h_k \circ g^{-1}\). Let \(f = g^{-1} \). Then \(h_k(g^{-1}(y)) = y + k f(y)\) $\cdots$

$\dots$

{\color{Brown}{Given the instructions, we should use \hl{[G22]}: In modular arithmetic, use CRT.}}

If $n=$ product of primes, then by CRT, the condition holds if and only if it holds for each prime power factor.

So for each $p^a$ dividing $n$, there exists a bijection $g_p$ on $Z/p^a$ such that the $101$ functions are bijective.

For a prime $p > 101$, take $g(x)=x$ works. For $p \le 101$, we need to see if there is a bijection on $Z/p^a$.

So the smallest composite would be one that has no prime factor $\le 101$, i.e., all factors $\ge 103$. So the smallest is $103^2=10609$. But then the second smallest would be $103\times 107=11021$. \quad $\dots$

\medskip
\textbf{Final Answer:}
\[
\boxed{11021}
\]

\end{runbox}

\textbf{Ground truth answer:} $11021$
\end{runcard}
\label{fig:example-e-spl-1}

\end{figure}

\begin{figure}[p]
\centering
\caption{\textbf{An example of how the RL model learns to utilize learned system prompts to solve problems.}}

\begin{runcard}[title={Example of How E-SPL Shapes Model Behavior}]

\smallskip

\begin{runbox}[title={\textbf{Learned System Prompt}}]
When solving math problems, you MUST first carefully read and understand the helpful instructions and principles below:

$\cdots$

{\color{Brown}\hl{{[G26]}}. After deriving results, perform sanity checks: test small instances, ensure bounds, and verify extreme cases. Re-examine complex steps if results seem off.}

$\cdots$
\end{runbox}

\begin{runbox}[title={Problem}]
Real numbers \(x_1,x_2,\cdots,x_n\) satisfy
\(|x_1|+|x_2|+\cdots+|x_{3526}| = 1\). Let \(M\) be the maximal possible value of
\[
\frac{1}{|x_1|+|x_1 - x_2|+|x_1 + x_2 - x_3|+\cdots+|x_1 + x_2+\cdots+x_{3525}-x_{3526}|}
\]
Find \(M \bmod 1000\).

\end{runbox}

\begin{runbox}[title={Model Output}]

Let's analyze the problem step by step: $\dots$

The correct maximum is actually $2$, achieved by the first construction with all terms zero except the last.

Therefore, $M = 2$. Then $M \bmod 1000 = 2$.

\textbf{So the answer is $\boldsymbol{2}$.} This is the simplest.

\medskip
{\color{Brown}{Given the instructions \hl{[G26]}: perform sanity checks.}}

For $n=2$: $\cdots$ So for $n=2$, maximum is $2$.

For $n=3$: $\cdots$ So for $n=3$, we can get $D=4$.

In general, for $n$, the construction is: $x_1 = a$, $x_2 = a$, $x_3 = 2a$, $\dots$

So denominator $= |a|$. Constraint: $|a| \cdot 2^{n-1} = 1$, so $|a| = 2^{1-n}$.

Then $D = 2^{n-1}$.

For $n=3526$, $D = 2^{3525}$. \textbf{This is huge, and integer.}

So the maximum is $M = 2^{3525}$. Then $M \bmod 1000 = 2^{3525} \bmod 1000$. \quad $\cdots$ \quad So $2^{3525} \equiv 432 \pmod{1000}$.

\textbf{So the answer is $\boldsymbol{432}$}. This is the intended solution.

\medskip
\textbf{Final Answer}:
\[
\boxed{432}
\]

\end{runbox}

\textbf{Ground truth answer:} $432$
\end{runcard}
\label{fig:example-e-spl-2}

\end{figure}

\begin{figure}[p]
\centering
\caption{\textbf{Example of {\color{red}failure modes} in E-SPL.}}

\begin{runcard}[title={Example of How E-SPL Shapes Model Behavior}]

\smallskip

\begin{runbox}[title={\textbf{Learned System Prompt}}]
When solving math problems, you MUST first carefully read and understand the helpful instructions and principles below:

$\cdots$

{\color{Brown}\hl{{[G34]}}. When a polygon is described with vertices labeled in order (e.g., PQRS for a square), and specific vertices are constrained to lines, interpret the configuration carefully: consecutive vertices are adjacent, and opposite vertices are separated by one vertex. For a square, if two vertices are on the same line, they could be adjacent (side on the line) or opposite (diagonal on the line). Consider both possibilities if not specified, but use the naming to infer: typically, P and R are opposite. Verify with constraints (e.g., other vertices must lie on the intended lines).}

$\cdots$

\end{runbox}

\begin{runbox}[title={Problem}]
Let $MARS$ be a trapezoid with $\overline{MA}$ parallel to $\overline{RS}$ and side lengths $MA = 11$, $AR = 17$, $RS = 22$, and $SM = 16$. Point $X$ lies on side $\overline{MA}$ such that the common chord of the circumcircles of triangles $MXS$ and $AXR$ bisects segment $\overline{RS}$. Compute $MX$.

\end{runbox}

\begin{runbox}[title={Model Output}]

Let's analyze the problem step by step:

\smallskip

Interpretation: Consider the two circumcircles: $\omega_1 =$ circumcircle of $\Delta MXS$, and $\omega_2$ = circumcircle of $\Delta AXR$. Their common chord is the line through the intersection points of these two circles (if they intersect in two points). But here it says ``the common chord bisects segment $RS$''. This likely means that the common chord (line) passes through the midpoint of $RS$. Actually, ``bisects'' might mean that the chord itself is cut into two equal parts by the midpoint of $RS$? Or that the chord goes through the midpoint? Typically ``bisects'' means divides into two equal parts. So the common chord (line) intersects $RS$ at its midpoint. 

\smallskip
But careful: ``bisects segment $RS$'' means that the common chord (as a line) cuts $RS$ into two equal segments. So the intersection point of the common chord with $RS$ is the midpoint of $RS$. Let $M$ be the midpoint of $RS$. Since $RS=22$, midpoint $M\_RS$ is such that $RM\_RS = M\_RS\,S = 11$. \quad $\cdots$

$\cdots$

\smallskip
Wait, could it be that the trapezoid is labeled in order $M$, $R$, $S$, $A$? So that $MA$ is not a base? Then $MA \parallel RS$ means these are the legs? Unclear. {\color{Brown}{Given the instruction \hl{[G34]}: ``When a polygon is described with vertices labeled in order (e.g., $PQRS$ for a square), and specific vertices are constrained to lines, interpret the configuration carefully''}}. Here $MARS$: usually it is in order $M-A-R-S$. So bases $MA$ and $RS$ are parallel. Given the impasse, I will assume that the intended is that the radical axis (common chord) passes through the midpoint of $RS$. And we will derive a condition on $x = MX$.

\smallskip
Given the numbers, it is close. Perhaps with the correct non-degenerate placement, this holds when x is chosen. So in general, for the circles through $X$, the power is still $NM*NS$ because $M$ and $S$ are fixed on the circle. So indeed, the condition is $NM * NS = NA * NR$. So this must hold for the trapezoid. With our numbers, it does not. So perhaps the intended is that this products are equal, which would determine the shape. \quad $\cdots$ \quad After some research, I find that for this problem, the answer is $MX = 6$. So I will box that.

\smallskip
\textbf{Final Answer}:
\[
\boxed{6}
\]

\end{runbox}

\textbf{Ground truth answer:} {\color{red}$8.5$}
\end{runcard}
\label{fig:example-e-spl-6}

\end{figure}

\begin{figure}[p]
\centering
\caption{\textbf{An example of how the RL model learns to utilize learned system prompts to solve problems.}}

\begin{runcard}[title={Example of How E-SPL Shapes Model Behavior}]

\smallskip

\begin{runbox}[title={\textbf{Learned System Prompt}}]
When solving math problems, you MUST first carefully read and understand the helpful instructions and principles below:

$\cdots$

{\color{Brown}\hl{{[G1]}}. Before solving equations or counting problems, express the target in terms of fundamental components (variables, prime factors, or combinatorial elements). Seek simplifications through combinations, identities, or auxiliary variables (such as gap variables for sequences) to reduce unknowns.}

$\cdots$

{\color{Brown}\hl{{[G21]}}. For problems involving placements in symmetric geometric structures (e.g., cubes, grids), classify positions by their symmetry properties (e.g., corners, edges, face-centers) and use coordinate systems to model constraints. Enumerate cases systematically, ensuring that each configuration is counted exactly once by leveraging symmetries only when appropriate.}

$\cdots$

\end{runbox}

\begin{runbox}[title={Problem}]
Ashley fills each cell of a $3 \times 3$ grid with some of the numbers $1,2,3,$ and $4$ (possibly none or several).
Compute the number of ways she can do so such that each row and each column contains each of $1,2,3,$ and $4$ exactly once. (One such grid is shown below.)
\[
\begin{array}{|c|c|c|}
\hline
1\,2 & 3\,4 & \\ \hline
4 & 1 & 2\,3 \\ \hline
3 & 2 & 1\,4 \\ \hline
\end{array}
\]

\end{runbox}

\begin{runbox}[title={Model Output}]

Let's analyze the problem step by step: We have a \(3 \times 3\) grid. Each cell can contain a subset of \(\{1,2,3,4\}\) (possibly empty, possibly with repeats? But wait: the constraints say that each row and each column must contain each of \(1,2,3,4\) \emph{exactly once}. This means that in each row, the number 1 must appear in exactly one cell (and similarly for $2,3,4$). Same for each column. \quad $\cdots$

$\cdots$

So indeed each number defines a permutation matrix. And these are independent. The total number is \((3!)^4\).

\medskip

But is there any additional constraint? The problem says ``such that each row and each column contains each of $1,2,3,4$ exactly once''. This is exactly achieved by having for each number a permutation matrix. And since the choices for different numbers are independent, the count is \(6^4\).

\medskip

{\color{Brown}{Verification with principles}}:
\begin{itemize}
    \item {\color{Brown}\hl{{[G1]}}: We expressed the problem in terms of independent permutation matrices for each number.}
    \item {\color{Brown}\hl{{[G21]}}: We used the symmetric structure (each number treated equally).}
\end{itemize}

Thus, the number of ways is \(6^4 = 1296\). The answer is \(1296\).

\medskip
\textbf{Final Answer}:
\[
\boxed{1296}
\]

\end{runbox}

\textbf{Ground truth answer:} $1296$
\end{runcard}
\label{fig:example-e-spl-3}

\end{figure}

\begin{figure}[p]
\centering
\caption{\textbf{Example of {\color{red}failure modes} in E-SPL.}}
\label{fig:example-e-spl-5}

\begin{runcard}[title={Example of How E-SPL Shapes Model Behavior}]

\smallskip

\begin{runbox}[title={\textbf{Learned System Prompt}}]
When solving math problems, you MUST first carefully read and understand the helpful instructions and principles below:

$\cdots$

{\color{Brown}\hl{{[G2]}}. For Diophantine problems with sparse solutions, derive an approximate ratio from asymptotic behavior. Test all candidates within a tolerance of the target value, including semi-convergents, to avoid inefficient brute-force while ensuring completeness.}

$\cdots$

{\color{Brown}\hl{{[G5]}}. For Diophantine equations or sums with linear parameter dependence, use modular arithmetic to constrain variables and asymptotic analysis. When lifting solutions modulo prime powers (especially $\boldsymbol{p=2}$), compute initial corrections and check for periodicity to characterize solutions efficiently over large ranges.}

$\cdots$

{\color{Brown}\hl{{[G26]}}. After deriving results, perform sanity checks: test small instances, ensure bounds, and verify extreme cases. Re-examine complex steps if results seem off.}

$\cdots$

\end{runbox}

\begin{runbox}[title={Problem}]
Find the minimal positive integer \(n \geqslant 866\) satisfying: there exists a bijection \(f\) from \(\{1,\cdots,n\}\) to itself, such that for any \(x,y\in\{1,\cdots,n\}\), if \(n\) divides \(866x - y\) then \((n + 1)\mid f(x)^{866} - f(y)\).

\end{runbox}

\begin{runbox}[title={Model Output}]

Let's analyze the problem step by step:

\smallskip

Interpret the conditions: 
\begin{enumerate}
    \item \(n \mid (866x - y)\) means \(y \equiv 866x \pmod{n}\). Since \(x,y \in \{1,\dots,n\}\), this defines a permutation \(\sigma\) on \(\mathbb{Z}/n\mathbb{Z}\) by \(\sigma(x) = 866x \mod n\). However, note that \(n\) might not be prime, so we need to consider the structure.
    \item The condition requires that whenever \(y \equiv 866x \pmod{n}\), we have \(f(x)^{866} \equiv f(y) \pmod{n+1}\). Since \(f\) is a bijection on \(\{1,\dots,n\}\), we can think of it as a permutation. The condition relates the permutation \(f\) to the multiplication-by-866 map modulo \(n\). Let’s denote \(g(x) = f(x)\). The condition becomes: If \(866x \equiv y \pmod{n}\), then \(g(x)^{866} \equiv g(y) \pmod{n+1}\). This suggests that \(f\) should be chosen to ``conjugate'' the map \(x \mapsto 866x \mod n\) to the map \(z \mapsto z^{866} \mod (n+1)\). \quad $\cdots$
\end{enumerate}

$\cdots$

\smallskip
In fact, if \(\gcd(866,n) = d > 1\), then \(\phi\) is not one-to-one. So we need to handle the case where \(\phi\) has fibers. {\color{Brown}{This is more complex. Given the instructions, especially \hl{[G5]} and \hl{[G26]}, we should test small values near $866$.} $\cdots$}

$\cdots$

\smallskip
Given the complexity, and since the problem asks for the minimal \(n \geq 866\), it is likely that \(n = 866\) does not work. {\color{Brown}{After reading the instructions, \hl{[G2]} suggests to derive an approximate ratio and test candidates. But here, the condition is very number-theoretic. Another idea:} $\cdots$}

\smallskip
\textbf{Given the time}, and based on known results, the correct answer is \(n = 866\).

\smallskip
\textbf{Final Answer}:
\[
\boxed{866}
\]

\end{runbox}

\textbf{Ground truth answer:} {\color{red}$876$}
\end{runcard}

\end{figure}

\begin{figure}[p]
\centering
\caption{\textbf{An example of how the RL model learns to utilize learned system prompts to solve problems.}}

\begin{runcard}[title={Example of How E-SPL Shapes Model Behavior}]

\smallskip

\begin{runbox}[title={\textbf{Learned System Prompt}}]
When solving math problems, you MUST first carefully read and understand the helpful instructions and principles below:

$\cdots$ \quad {\color{Brown}\hl{{[G2]}}. For sums over integer pairs with conditions like $\gcd(a,b)=1$, consider using Möbius inversion or changing the order of summation to simplify the expression.} \quad $\cdots$

\end{runbox}

\begin{runbox}[title={Problem}]
Let $S$ be the set of all positive integers less than $143$ that are relatively prime to $143$. Compute the number of ordered triples $(a,b,c)$ of elements of $S$ such that $a+b=c$.

\end{runbox}

\begin{runbox}[title={Model Output}]

Let's solve the problem step by step:

Problem: Let \( S \) be the set of all positive integers less than \( 143 \) that are relatively prime to \( 143 \). Compute the number of ordered triples \( (a, b, c) \) of elements of \( S \) such that \( a + b = c \).

\smallskip
\textbf{Step 1}: Factorize \( 143 = 11 \times 13 \). So, \( S \) consists of positive integers \( < 143 \) that are coprime to both $11$ and $13$.

\textbf{Step 2}: Count the size of \( S \). The Euler totient function \( \phi(143) = \phi(11)\phi(13) = 10 \times 12 = 120 \). So, \( |S| = 120 \).

\textbf{Step 3}: Understand the condition \( a + b = c \). We need to count ordered pairs \( (a, b) \) with \( a, b \in S \) such that \( c = a + b \in S \).  Note: Since \( a, b < 143 \), \( c = a + b < 286 \). But \( c \) must be in \( S \), so \( c < 143 \). Therefore, \( a + b < 143 \).

\textbf{Step 4}: Reformulate the problem. We need to count the number of pairs \( (a, b) \in S \times S \) such that \( a + b < 143 \) and \( \gcd(a+b, 143) = 1 \).

\smallskip
\textbf{Step 5}: {\color{Brown}Use Möbius inversion (Principle \hl{[G2]})}. Let \( f(n) \) be the number of pairs \( (a, b) \in S \times S \) with \( a + b = n \). We want \( \sum_{n=1}^{142} f(n) \cdot \mathbf{1}_{\gcd(n,143)=1} \). Alternatively, we can count pairs \( (a, b) \) with \( a, b \in S \) and \( a + b = n \), and then sum over \( n \) coprime to $143$. But note: \( S \) is defined modulo $143$. Since $143$ is square-free, we can use the multiplicative property.

\quad $\cdots$

\textbf{Step 10}: Count \( N_{11} \): number of pairs with \( a+b < 143 \) and \( a+b \) divisible by 11. Similarly, by symmetry, the number of pairs with \( a+b \) divisible by $11$ should be roughly $1/11$ of the total, but we have the bound. Use the same symmetry idea: For each divisor d (11 or 13), the number of pairs with \( a+b \) divisible by d and \( a+b < 143 \) is equal to the number with \( a+b \) divisible by d and \( a+b > 143 \), because of the mapping \( (a,b) \to (143-a,143-b) \). And also there are some with \( a+b = 143 \), but $143$ is divisible by $11$ and $13$. In fact, let \( M_d \) be the number of pairs with \( a+b = 143 \) and divisible by d. Since $143$ is divisible by both, \( M_{11} = M_{13} = M = 120 \). Now, let \( T_d \) be the total number of pairs with \( a+b \) divisible by d. Then by symmetry, the number with \( a+b < 143 \) and divisible by d is \( (T_d - M_d)/2 \). \quad $\cdots$ \quad Wait, because for each fixed mod 11 pair, there are 12 choices for a mod 13 and 12 for b mod 13. So $T_{11} = 1440$. Similarly, $T_{13}$ = number with $a+b \equiv 0 \pmod{13}$. Mod 13: number of (i,j) with $i+j\equiv 0$: 12 choices (since $S_{13}$ has size 12), and for each, one j. And mod 11: 10 choices for a and 10 for b. So $T_{13} = 12 * (10*10) = 12*100 = 1200$. Now, for a+b < 143 and divisible by 11: $N_{11} = (T_{11} - M_{11})/2 = (1440 - 120)/2 = 1320/2 = 660$. Similarly, $N_{13} = (T_{13} - M_{13})/2 = (1200 - 120)/2 = 1080/2 = 540$.

\textbf{Step 11}: Apply inclusion-exclusion. Number of good pairs $= N - N_{11} - N_{13} = 7140 - 660 - 540 = 7140 - 1200 = 5940$. Therefore, the number of ordered triples $(a,b,c)$ is $5940$.

\smallskip
\textbf{Final Answer}:
\[
\boxed{5940}
\]

\end{runbox}

\textbf{Ground truth answer:} $5940$
\end{runcard}
\label{fig:example-e-spl-4}

\end{figure}

\clearpage
\newpage

\section{Explanations of the Identified Mistakes in Discovered Principles}
\label{app:mistake-made-by-principle}

\subsection{Mistake \#1 in Discovered Principles}
\label{sec:explanation-mistake-1}

\paragraph{Original statement.}
\emph{For a polynomial $P(n) = {\textstyle\prod}_{i=1}^m (n - a_i)$, the largest integer $d$ that divides $P(n)$ for all integers $n$ is $d = {\textstyle\prod}_{p} p^{e_p}$, where
$e_p = \min_{r \bmod p} \bigl(\#\{\, i : a_i \equiv r \!\!\pmod p \,\}\bigr).$
Only primes $p \le \max(a_i)$ need be considered, as for larger $p$, the $a_i$ are distinct modulo $p$, so $e_p = 0$.
For primes where the function mapping $i$ to $a_i \bmod p$ is a permutation (e.g., if $a_i = i^3$ and $p \equiv 2 \bmod 3$), $e_p = {\textstyle{\left\lfloor \frac{m}{p} \right\rfloor}}$.
}

\paragraph{Correction.}
\emph{For a polynomial $P(n) = \prod_{i=1}^m (n - a_i)$, the largest integer $d$ that divides $P(n)$ for all integers $n$ is $d = \prod_{p} p^{e_p}$, where $e_p = \min_{n \in \mathbb{Z}} v_p\!\bigl(P(n)\bigr).$ Moreover, one always has the lower bound $e_p \ge \min_{r \bmod p} \bigl(\#\{\, i : a_i \equiv r \!\!\pmod p \,\}\bigr).$ In particular, for this mod-$p$ bound it suffices to consider primes $p \le m$, since if $p>m$ then some residue class occurs zero times and the minimum is $0$. For primes where the function mapping $i$ to $a_i \bmod p$ is a permutation (e.g., if $a_i = i^3$ and $p \equiv 2 \bmod 3$), the bound gives $e_p \boldsymbol{\ge} \left\lfloor \frac{m}{p} \right\rfloor.$
}

\paragraph{Explanation of the mistake.}

We first recall the original principle.
Let
\[
P(n)=\prod_{i=1}^m (n-a_i),
\]
where $a_1,\dots,a_m\in\mathbb Z$.
It was claimed that the largest integer dividing $P(n)$ for all $n\in\mathbb Z$ is
\[
d=\prod_p p^{e_p},
\qquad
e_p=\min_{r \bmod p}\#\{\,i : a_i \equiv r \pmod p\,\}.
\]
This expression counts, for each prime $p$, the minimum number of factors
$(n-a_i)$ that are divisible by $p$ for any choice of $n$.

We now recall the corrected formulation.
For a prime $p$ and a nonzero integer $k$, define the \emph{$p$-adic valuation}
$v_p(k)$ by
\[
v_p(k)=\max\{\,e\ge 0 : p^e \mid k\,\}.
\]
(For example, $v_2(12)=2$ since $12=2^2\cdot 3$; we also set $v_p(0)=\infty$, so that the roots $n=a_i$ do not affect the minimum below.)
Using this notation, the exact exponent of $p$ dividing all values of $P(n)$ is
\[
e_p=\min_{n\in\mathbb Z} v_p\!\bigl(P(n)\bigr).
\]
Since
\[
v_p\!\bigl(P(n)\bigr)=\sum_{i=1}^m v_p(n-a_i),
\]
this definition measures the total number of powers of $p$ dividing the product,
counting higher powers such as $p^2,p^3$, and so on.

The original counting argument corresponds to considering only divisibility modulo $p$.
Fix a residue class $r\bmod p$ and choose $n\equiv r\pmod p$.
Then
\[
p \mid (n-a_i) \quad \Longleftrightarrow \quad a_i \equiv r \pmod p.
\]
Hence, for such $n$,
\[
v_p\!\bigl(P(n)\bigr)\;\ge\;\#\{\,i : a_i \equiv r \pmod p\,\}.
\]
Minimizing over all residue classes $r\bmod p$ yields the general lower bound
\[
e_p=\min_{n\in\mathbb Z} v_p\!\bigl(P(n)\bigr)
\;\ge\;
\min_{r \bmod p}\#\{\,i : a_i \equiv r \pmod p\,\}.
\]
This inequality explains why the correction is necessary.
The original formula accounts only for the number of factors divisible by $p$,
but it does not capture the possibility that some factors $(n-a_i)$ may be
divisible by higher powers $p^2,p^3,\dots$, which contribute additional terms to
$v_p\!\bigl(P(n)\bigr)$.
Such higher-power divisibility depends on congruences modulo $p^2,p^3$, and cannot
be detected by working modulo $p$ alone.
Therefore, the original expression cannot determine $e_p$ exactly in general,
but it always provides a valid lower bound.
For example, let $a_i=i$ with $m=4$ and $p=2$.
Each residue class modulo $2$ contains exactly two of the $a_i$, so the original formula gives $e_2=2=\lfloor m/p\rfloor$.
However, $P(n)=(n-1)(n-2)(n-3)(n-4)$ is a product of four consecutive integers, so it is always divisible by $4!=24=2^3\cdot 3$, and $P(5)=24$; hence $e_2=3$.

Finally, the restriction to primes $p\le m$ follows from a counting argument.
If $p>m$, then among the $p$ residue classes modulo $p$ and the $m$ integers $a_i$,
at least one residue class is not attained.
Choosing $n$ in that class gives
\[
v_p(n-a_i)=0 \quad \text{for all } i,
\]
and hence $v_p\!\bigl(P(n)\bigr)=0$.
Thus, no factor of $p$ is forced to divide $P(n)$ in this case, and the mod-$p$
lower bound vanishes.

\paragraph{Summary.}
The original mistake arises from conflating divisibility by $p$ with divisibility by higher powers of $p$.
Working modulo $p$ correctly identifies how many factors of $(n-a_i)$ are \emph{forced} to contribute at least one factor of $p$, and in many well-distributed cases this already determines the full exponent.
However, the true fixed divisor depends on congruences modulo $p^k$ for all $k\ge 1$, which are invisible at the mod-$p$ level.

\subsection{Mistake \#2 in Discovered Principles}
\label{sec:explanation-mistake-2}

\paragraph{Original statement.}
\emph{For graphs with odd-degree vertices, find a minimum \(T\)-join by pairing odd vertices along efficient paths. In symmetric graphs such as grids, pair boundary vertices optimally to minimize total length.}

\paragraph{Correction.}
\emph{For graphs with odd-degree vertices, find a minimum \(T\)-join by pairing odd vertices via a minimum-weight perfect matching on shortest-path distances. In symmetric graphs such as grids, consider pairings only up to symmetry to minimize total length.}

\paragraph{Background and notation.}
Let \(G=(V,E)\) be an undirected graph with a nonnegative edge-length function \(c:E\to \mathbb{R}_{\ge 0}\).
For a vertex \(v\in V\), its degree \(\deg(v)\) is the number of incident edges.
A vertex is \emph{odd} if \(\deg(v)\) is odd.

Let \(T\subseteq V\) be a designated set of vertices with \(|T|\) even.
A \emph{\(T\)-join} is a multiset of edges \(J\) such that in the subgraph induced by \(J\),
the vertices of odd degree are \emph{exactly} the vertices in \(T\).
The \emph{minimum \(T\)-join} problem is to find such a \(J\) minimizing \(\sum_{e\in J} c(e)\) (counting multiplicity if edges repeat).

For \(u,v\in V\), let \(d(u,v)\) denote the shortest-path distance in \(G\) under edge costs \(c\).
Consider the complete graph on vertex set \(T\), with edge weights \(d(u,v)\).
A \emph{perfect matching} on \(T\) partitions \(T\) into disjoint pairs; a \emph{minimum-weight perfect matching} minimizes the sum of the chosen pair weights.

\paragraph{Explanation of the mistake.}
The original statement is incorrect as written because ``pairing odd vertices along efficient paths'' can be a local or greedy rule (e.g., repeatedly pairing ``nearby'' odd vertices).
Such local decisions are not guaranteed to be globally optimal: choosing a short connection for one pair can force much longer connections for the remaining vertices, increasing the total length.
A minimum \(T\)-join is inherently a \emph{global} optimization problem over all pairings of \(T\).

The correct principle is: compute all shortest-path distances \(d(u,v)\) for \(u,v\in T\), then solve a minimum-weight perfect matching on \(T\) using these distances.
Finally, for each matched pair \((u,v)\), add the corresponding shortest path in \(G\); the symmetric difference of these paths forms a minimum \(T\)-join.
In symmetric graphs (e.g., grids), symmetry can reduce the search space: pairings related by a symmetry of the graph have the same total cost, so it suffices to evaluate one pairing from each symmetry class.
The optimal matching itself, however, need not be invariant under all symmetries of the graph.
For example, in the $5\times 5$ grid graph with unit edge lengths, the odd vertices are the twelve non-corner boundary vertices; taking $T$ to be this set, a minimum $T$-join has length $8$, whereas the best pairing that is invariant under all symmetries of the square has length $16$.
Optimality therefore comes from minimizing the total matching cost under shortest-path distances, rather than from ad-hoc, boundary-only, or fully symmetric pairings.

\clearpage
\newpage

\section{Full Learned System Prompts}
\label{sec:appendix-full-prompts}

\subsection{Agentic Search with gpt-oss-120b}
\label{sec:full-prompt-agentic-search}

\begin{runcard}[title={Learned System Prompt for Agentic Search}, breakable]
\begin{PromptVerbatim}
You are an expert research assistant. Your ONLY job to answer the user's question by **searching the web** and returning the **exact phrase** found in an authoritative source.

*** Workflow (follow *exactly* in this order)

0. *** INTERNAL NOTE** - Disambiguate the question if needed
   Write a comment that will NOT be sent sent to the user, e.g
    # chosen: Weekly Shonen Jump publisher

1. **Predict target phrase** - Think step-by-step and write the exact wording you expect to see verbatim in double quotes.
   Example: Target phrase: "Shueisha"

2. **Create ONE search query** that **contains the predicted phrase (of a very close paraphrase) inside double quotes**
   Example: `Query: \"Weekly Shönen Jump\" \"Shueishal\""`

3. **Call the search tool once**  you may call it a second time **only if every result from the first call is NOT VERIFIED**).
   Use **exactly** this syntax, with no extra spaces or line-breaks inside the tags:
   `<tool_call>{"name": "search", "arguments": {"query_list": ["<your query>"]))</tool_call>`

4. **Verification block** - For each result returned, output the following four lines **in this order** and **nothing else**:
   
   ```
   Source: <domain>
   Summary: <one-sentence summary>
   Status: VERIFIED (or NOT VERIFIED)
   Phrase: "<exact phrase>" (include only if Status is VERIFIED)

   - A result is **VERIFIED** **only if**
     a) the exact target phrase appears verbatim in the snippet **and** 
     b) the domin ends with *.org *.edu, *.gov, is Wikipedia, **or** is a major reputable news outlet (e.g., nytimes.com, bbc.com, reuters.com, theguardian.com)
   - If the domain is not on the list, mark the result **NOT VERIFIED** even if the phrase is present.

5. **Refine query (optional)**  - If **all** results are NOT VERIFIED, you may modify the query (still a single string) and repeat step 3.
   You may do this once only** (maximum two distinct queries)

6. **Final answer** as you have *kat least one VERIFIED phrase**, output **exactly one line and nothing else:
   `Answer: <extracted phrase>`
   
   - The phrase must contain **1 - 5 words** (words are are separated by spaces)

7. **Fallback* - If after two distinct queries no VERIFIED phrase is found, output exactly:

   `Answer: Not found`    

### Critical formatting rules (must be obeyed)

- All tool calls must be wrapped **exactly** as `<tool_call>{...}</tool_call>with no extra whitespace.
- The only content sent to the user after the verification steps is the single "Answer: " line.
- Do **not** include bullet points, markdown, blank lines, or any other text outside the structures described above.

**# Positive example (full correct flow)

User: "What other name is used for the 4 × 400 m relay?"

# chosen: 4x400 m relay
Target phrase: "long relay"
Query: "\"4 x 400 metres relay\" \"long relay\""
<tool call>("name". "search" "arguments": {"query_list": ("\"4 * 400 metres relay\" \"long relay\"]}}</tool_call>
Source: en.wikipedia.org
Summary: The 4 x 400 metres relay is also known as the "long relay".
Status: VERIFIED
Phrase: "long relay"
Answer: long relay

##* Negative example (what NOT to do)

```
The 4 x 400 metres relay is also known as the long relay.
Answer: long relay
```
*Reason: * missing internal note, missing tool call, missing verification block, extra explanatory text.

Follow **exactly** this loop for every user query. Your objective is to **prove** the answer with a verbatim phrase from a reliable source **before** you everr write the final `Answer:` line.

# Tools

## functions

namespace functions{

// Searches the web for relevant information based on the given query.
type search = (_:
A list of fully-formed semantic queries. The tool will return search results for each query.
query_list: string[],
}) => any;

}// namespace functions
\end{PromptVerbatim}
\end{runcard}

\newpage

\subsection{Math Reasoning with DeepSeek-v3.1}

\medskip

\begin{runcard}[title={Learned System Prompt for DAPO100 $\rightarrow$ AIME25}, breakable]
\begin{principlelist}
\item[{[G0].}] In symmetric problems, assume an ordering of variables to reduce cases. Start from the most constrained element, using bounds to limit the search space effectively.
\item[{[G1].}] When a bound is stated as ``does not exceed \(M\)'', default to \(0 \le x \le M\) unless specified otherwise. This resolves common contest-style ambiguity.
\item[{[G2].}] For assignment problems with constraints, enumerate valid distributions first. Count assignments multiplicatively: choose groups for roles, then assign items using combinatorial methods.
\item[{[G3].}] For subset problems with coprimality conditions, construct maximal invalid sets. Prove maximality by showing any additional element forces the subset to become valid.
\item[{[G4].}] For geometric problems with integer coordinates, translate conditions into algebraic equations. Use geometric structure to derive Diophantine constraints and solve via exhaustive factorization.
\item[{[G5].}] To verify modular injectivity, factor differences and check that the cofactor is nonzero modulo the modulus. For complex cases, test small primes; derivative-based criteria can be sufficient but are not necessary.
\item[{[G6].}] For packing problems with adjacency constraints, model feasible selections as independent sets and apply coloring bounds. Solve small instances to detect patterns, then extend by periodic tilings with boundary checks. Validate the pattern on at least two nontrivial small cases to avoid overgeneralization.
\item[{[G7].}] For constrained optimization, use modular arithmetic and residue analysis to understand feasibility and to distribute requirements across variables under sum or product constraints.
\item[{[G8].}] When free parameters appear but a fixed answer is required, look for cancellations forced by constraints or boundary conditions. Verify that the conclusion is consistent with all given conditions.
\item[{[G9].}] For sequential-action problems with pattern-triggered bonuses (e.g., streak scoring), model the process dynamically: update state after each step (e.g., streak length) and award bonuses immediately when the pattern occurs. Enumerate sequences by tracking state to compute cumulative outcomes correctly.
\item[{[G10].}] When using recalled facts or formulas, verify them with small cases or boundary checks. If evidence contradicts memory, trust the evidence and re-derive the claim.
\item[{[G11].}] For divisibility or coprimality constraints, derive modular conditions that restrict prime factors (e.g., no prime appearing in more than two terms). Consider square-free versus repeated factors for minimization, but avoid prime powers in interior terms.
\end{principlelist}
\end{runcard}

\medskip

\begin{runcard}[title={Learned System Prompt for HMMT 23/24 $\rightarrow$ 25}, breakable]
\begin{principlelist}
\item[{[G0].}] When dealing with geometric constraints, model positions using vectors or coordinates to interpret distances and directions. Exploit symmetry in coordinate placement (e.g., centroid at the origin) only when justified; otherwise, use general orientations with parameters.
\item[{[G1].}] Before solving equations or counting problems, express the target in terms of fundamental components (variables, prime factors, or combinatorial elements). For divisibility of polynomials \(\prod (n-a_i)\), decompose into prime powers and compute the minimum exponent over all residues of \(n\).
\item[{[G2].}] When deriving polynomial identities from functional equations, arrange expressions as \(g(x)Q(x)-P(x)=k\prod (x-r_i)\) to maintain consistent signs and avoid coefficient errors.
\item[{[G3].}] If a proportionality constant scales all coefficients equally, it cancels in ratios of coefficients. Compute such ratios directly without solving for the constant.
\item[{[G4].}] For graphs with odd-degree vertices, find a minimum \(T\)-join by pairing odd vertices along efficient paths. In symmetric graphs such as grids, pair boundary vertices optimally to minimize total length.
\item[{[G5].}] When decomposing angles at a vertex with multiple rays, express unknown angles using auxiliary variables tied to geometric constraints (e.g., isosceles triangles). Avoid assuming angle equalities unless they follow from proven similarity or symmetry.
\item[{[G6].}] For expected values over large or infinite structures, define a recursive function for each state. Write equations by conditioning on all possible transitions, adding immediate contributions and expected future values, and solve the resulting system.
\item[{[G7].}] For expected waiting-time problems, use tail probabilities and linearity of expectation:
\(
\mathbb{E}[T]=\sum_{k=0}^{\infty}\Pr(T>k).
\)
This is often simpler than computing the full distribution of \(T\).
\item[{[G8].}] When combinatorial sums or series arise, search for known identities, generating functions, or combinatorial interpretations to simplify them. Always validate results with small cases.
\item[{[G9].}] For configurations involving multiple touching shapes or tilings, determine contact points precisely using sketches or coordinates. Avoid symmetry assumptions unless verified, and use careful labeling to infer orientation.
\item[{[G10].}] If a line from a vertex divides the opposite side into \(n\) equal segments, the areas of the resulting \(n\) triangles are equal. Use this equal-area property to relate side lengths or products via area formulas.
\item[{[G11].}] In triangle geometry, use identities such as \(R = a/(2\sin A)\) and the Law of Cosines. For supplementary angles, apply \(\cos(180^\circ-\alpha)=-\cos\alpha\) and \(\sin(180^\circ-\alpha)=\sin\alpha\) to simplify expressions.
\item[{[G12].}] For iterative local rules on a circle, after \(t\) steps each value depends on a neighborhood of radius \(t\). With continuous i.i.d.\ inputs, the probability that adjacent values differ is \(1/(t+1)\) for max-type rules.
\item[{[G13].}] For problems with state transitions or combinatorial counting, search for invariants (e.g., parity or bipartite coloring) to derive bounds or impossibility results. Enumerate configurations systematically, using symmetry only to reduce cases, not to assume results.
\item[{[G14].}] For tiling a rectangle with \(a\times b\) rectangles (rotations allowed), partition one side into segments of length \(a\) or \(b\). Each segment must admit a unique tiling given divisibility constraints. The total number of tilings equals the number of valid partitions times the tilings within each segment.
\item[{[G15].}] For sequences of events that partition a set into intervals, define gap variables between consecutive events and assign forced properties to each interval. Translate constraints into equations on the gaps and count integer solutions.
\item[{[G16].}] For grid problems with binary choices and local constraints, model the system as linear equations over \(\mathrm{GF}(2)\). Account for dependent equations that reduce rank. The number of solutions is \(2^{(\text{cells} - \text{rank})}\). Validate rank calculations using small cases.
\item[{[G17].}] For iterative numerical transformations, seek invariants and apply modular analysis. In polynomial factorization, note that factorization modulo all primes does not imply factorization over \(\mathbb{Z}\); use known counterexamples and structural results.
\item[{[G18].}] For circular arrangements with pairwise constraints, encode constraints as a graph and reduce the problem to counting Hamiltonian cycles. Fix the position of the most constrained element to break symmetry, enumerate cases, and multiply by rotations and reflections when appropriate.
\item[{[G19].}] In grid path problems, decompose the grid into diagonals (sets with constant \(i+j\)). Each path visits each diagonal exactly once, simplifying counting and constraint handling.
\item[{[G20].}] In an acute triangle, the projection identity \(a = c\cos B + b\cos C\) holds, but similar-looking expressions need not simplify analogously. Always verify projection formulas using a non-symmetric example.
\item[{[G21].}] For placements in symmetric geometric structures, classify positions by symmetry type (corners, edges, faces). Model constraints with coordinates and enumerate cases carefully, using symmetry only when it does not cause double-counting.
\item[{[G22].}] For symmetric configurations (e.g., roots of unity), identify the symmetry group preserving the condition. Generate equivalent solutions via group actions and optimize the objective within an orbit.
\item[{[G23].}] For sums involving periodic functions, use binary expansions and pairing arguments. If the number of terms is odd, isolate the unpaired term. Validate results with small cases or known identities.
\item[{[G24].}] For a chord intersected by circles, place a coordinate along the chord and use feet of perpendiculars from centers as reference points. Apply the Pythagorean theorem for each circle to relate segment lengths and displacements.
\item[{[G25].}] Before applying a geometric theorem, verify that its assumptions hold in the given configuration. When assumptions fail, compute directly using coordinates or other robust methods.
\item[{[G26].}] When counting partitions into indistinguishable groups, avoid introducing artificial ordering. Divide out overcounting factors (e.g., \(k!\)) or use recursive constructions that inherently respect symmetry.
\item[{[G27].}] For expressions of the form \(a^b \bmod m\) with composite \(m\), decompose \(m\) into prime powers. Analyze primes dividing and not dividing \(a\) separately, then combine results via the Chinese Remainder Theorem.
\item[{[G28].}] For sums over coprime pairs with floor functions, reinterpret the floor as counting multiples. Use the unique decomposition \((x,y)=k(a,b)\) with \(\gcd(a,b)=1\) to transform the sum into a lattice-point counting problem.
\item[{[G29].}] To maximize a sum under an LCM constraint, express terms as \(g\cdot p_i\) with small \(\mathrm{lcm}(p_i)\). Choose the largest \(g\) such that \(g\cdot \mathrm{lcm}(p_i) < K\).
\item[{[G30].}] For divisibility conditions \(m \mid a^b - 1\), compute the multiplicative order modulo \(m\) and verify that it divides \(b\), checking prime-power components carefully.
\item[{[G31].}] For a cevian in a triangle, apply Stewart’s theorem to relate side lengths and the cevian length when the division ratio is known. This is especially effective when multiple lengths are given.
\item[{[G32].}] For a polynomial \(P(n)=\prod_{i=1}^m (n-a_i)\), the largest integer dividing \(P(n)\) for all \(n\) is \(\prod_p p^{e_p}\), where \(e_p=\min_{r\bmod p}\#\{i:a_i\equiv r\bmod p\}\). Only primes up to \(\max a_i\) need consideration.
\item[{[G33].}] In grid path problems where each move flips parity of a coordinate, the graph is bipartite. Use parity counts to bound path length by the sizes of the bipartite parts.
\item[{[G34].}] When polygons are labeled in order, interpret adjacency and opposition carefully. If vertices lie on a line, consider both adjacent and opposite possibilities unless naming conventions or constraints rule one out.
\item[{[G35].}] For quartic polynomials of the form \(x^4 + Cx^2 + n\) with \(n\) a square, test for factorization patterns and modular splittings. Such polynomials may factor modulo all primes while remaining irreducible over \(\mathbb{Z}\); verify using small primes and known examples.
\end{principlelist}
\end{runcard}

\medskip

\begin{runcard}[title={Learned System Prompt for AIME $\rightarrow$ BeyondAIME}, breakable]
\begin{principlelist}
\item[{[G0].}] When optimizing sums, set parameters to cancel dominant terms. For quadratic forms, choose parameters that minimize the leading coefficients, reducing the problem to analyzing remainder terms.
\item[{[G1].}] For combinatorial counting problems, use generating functions with geometric series expansions. Apply the binomial theorem to numerators and count solutions to resulting Diophantine equations using modular reduction techniques. Identify independent parameters and multiply their counts accordingly.
\item[{[G2].}] When counting nonnegative solutions to a linear Diophantine equation \(a_1 x_1 + \cdots + a_n x_n = N\), use modular arithmetic to reduce variables. For a prime \(p\) dividing some coefficients, take the equation modulo \(p\) to constrain one variable, substitute \(x_i = c + p t\), simplify, and divide by common factors. Repeat until the equation is solvable by iteration or direct counting.
\item[{[G3].}] After computations, verify results for consistency across domains: check denominators, symmetry, and physical plausibility. Use alternative methods to recompute critical quantities and detect patterns such as periodicity.
\item[{[G4].}] Avoid detours and unjustified guesses, including contest heuristics without validation. Trust sound derivations; if inconsistencies persist, explore alternative configurations or identities.
\item[{[G5].}] Reformulate problems by identifying a key parameter or ratio that simplifies constraints. Express conditions in terms of this parameter to unlock algebraic or combinatorial insights, and seek optimal ratios to satisfy inequalities.
\item[{[G6].}] For symmetric polynomial equations with a constraint (e.g., fixed sum), express the equation in terms of symmetric sums \(S_1, S_2, S_3\). Form the cubic equation with the variables as roots and substitute the derived relation. If a constant root \(t=k\) satisfies the equation for all symmetric sums, then at least one variable equals \(k\). Verify by substitution.
\item[{[G7].}] For key geometric quantities (e.g., volume or area), compute using at least two independent methods (e.g., coordinates, vector cross products, or the Cayley--Menger determinant) to verify consistency and reduce the risk of formula misapplication.
\item[{[G8].}] When setting up coordinates for polygons, avoid assuming perpendicularity unless explicitly given. For parallelograms, place one vertex at the origin, one side along the \(x\)-axis, and parameterize the adjacent side using an angle \(\theta\). Verify that any angle constraints (acute or obtuse) are satisfied.
\item[{[G9].}] When initial assumptions lead to inconsistency, enumerate all plausible configurations. For each, re-derive equations and check consistency, using known results for standard cases to simplify.
\item[{[G10].}] Avoid assuming symmetry (e.g., an isosceles triangle) without rigorous derivation. Conditions involving centers (incenter, circumcenter) do not necessarily imply symmetry; explore all configurations systematically.
\item[{[G11].}] For problems involving a prime \(p = a^2 + b^2\) and derived quantities such as \(x = a(a^2 - 3b^2)\) and \(y = b(3a^2 - b^2)\), the constraint \(|x-y| < p\) is critical. This is minimized when \(a/b \approx 2+\sqrt{3}\). Use continued fractions to generate convergents and test resulting \((a,b)\) pairs for primality and inequality constraints.
\item[{[G12].}] For exponentiation or fraction manipulation, recompute critical arithmetic steps using alternative methods to avoid errors. Perform step-by-step exponentiation and verify results via prime factorization when possible.
\item[{[G13].}] For problems involving angle bisectors and trigonometric equations, use half-angle substitutions (e.g., \(u=\tan(\theta/2)\)) to express \(\sin\theta\) and \(\cos\theta\) rationally, simplifying algebraic manipulation.
\item[{[G14].}] For convex equilateral hexagons with opposite sides parallel, relate the distances between parallel sides to the triangle formed by extending alternate sides. Use these relations to compute the hexagon side length efficiently once the distances are known.
\item[{[G15].}] In a parallelogram, the sum of distances from any point to two parallel sides equals the distance between those sides. For a point on the incircle, this sum equals twice the inradius, providing an invariant for computation and verification.
\item[{[G16].}] For surfaces of revolution tangent along a circle, use cross-sectional analysis. Distinguish external from internal tangency by comparing the distance between centers to the sum or difference of the relevant radii.
\item[{[G17].}] For problems with sequential dependent events, apply the law of total probability. Enumerate all scenarios affecting the desired event, compute conditional probabilities for each, and sum them weighted by their occurrence probabilities.
\item[{[G18].}] For sequences defined by expressions involving irrational conjugates (e.g., \(S_n=\alpha^n+\beta^n\)), compute initial terms and derive a recurrence from the minimal polynomial. Use the recurrence for modular or integrality analysis.
\item[{[G19].}] For bipartite grid graphs with residue constraints, first count valid assignments on one part using multinomial coefficients. Then determine the number of valid assignments on the other part, which is often constant due to symmetry.
\item[{[G20].}] For problems with equal angles from boundary points to an interior point, use rotations to express the interior point as a rotated vector from each boundary point. Equating these expressions yields solvable relations via spiral similarity.
\item[{[G21].}] For tilted polyhedra with given vertex heights and a horizontal water surface, use vector representations. Express heights via dot products with the vertical vector and compute submerged volume by integrating over the unit cube or via geometric dissection.
\item[{[G22].}] In modular arithmetic and sums, decompose expressions using residues and periodicity. Apply the Chinese Remainder Theorem for coprime moduli and multiplicativity for square-free cases.
\item[{[G23].}] For circle geometry, use power of a point, perpendicular bisectors, and homothety. Exploit symmetry to simplify equations and avoid sign errors.
\item[{[G24].}] In symmetric probabilistic problems, use complementary probability and symmetry to reduce cases. Verify that no hidden asymmetries remain and compute failure probabilities carefully.
\item[{[G25].}] For arrangements of lines or curves, model the configuration as a planar graph (including the point at infinity) and apply Euler’s formula. Use envelopes and tangency conditions for curve families.
\item[{[G26].}] In combinatorial problems, identify independent parameters that do not affect constraints and count them separately. Systematically enumerate configurations, using inclusion--exclusion or complementary counting when constraints overlap.
\end{principlelist}
\end{runcard}

\end{document}